\documentclass[a4paper,11pt]{article}
\usepackage[top=4cm, bottom=4cm, left=3cm, right=3cm]{geometry}
\usepackage{amssymb}
\usepackage{amsmath}
\usepackage{amsthm}
\usepackage{mathrsfs} 
\usepackage[usenames,dvipsnames]{pstricks}
\usepackage{euler}
\usepackage{euscript}
\usepackage{graphicx}
\usepackage{charter}
\usepackage{mdframed}
\usepackage{enumerate, subfigure, color}
\usepackage[colorlinks,hyperindex]{hyperref}
\usepackage{bibunits}

\usepackage{newfloat}
\usepackage{caption}

\usepackage{algorithmicx}
\usepackage{algpseudocode} 
\DeclareFloatingEnvironment[fileext=frm,placement={!ht},name=Algorithm]{algorithm}

\usepackage{booktabs}
\usepackage{cite}

\title{\sffamily\LARGE Less is More: \Nystrom{} Computational Regularization}
\author{Alessandro Rudi $^{1}\qquad$ Raffaello Camoriano $^{1,2}\qquad$ Lorenzo Rosasco $~^{1,3}$\\
{\small{\em ale\_rudi@mit.edu}$~~\qquad\qquad${\em raffaello.camoriano@iit.it}$\qquad\qquad$ {\em lrosasco@mit.edu}}$~\quad$\\[5mm]
{\small ${}^1$ Universit\`a degli Studi di Genova - DIBRIS, Via Dodecaneso 35, Genova, Italy}\\
{\small ${}^2$ Istituto Italiano di Tecnologia - iCub facility, Via Morego 30, Genova, Italy}\\
{\small ${}^3$ Massachusetts Institute of Technology and Istituto Italiano di Tecnologia -}\\ {\small Laboratory for Computational and Statistical Learning,}
\\ {\small Cambridge, MA 02139, USA}
}

\newcommand{\Nystrom}[1]{{Nystr\"om}}

\providecommand{\scal}[2]{\left\langle{#1},{#2}\right\rangle}
\providecommand{\nor}[1]{\lVert{#1}\rVert}
\providecommand{\ran}[1]{\operatorname{ran}{#1}}
\providecommand{\tr}{\operatorname{Tr}}

\newcommand{\R}{\mathbb R}
\newcommand{\C}{\mathbb C}
\newcommand{\N}{\mathbb N}

\newcommand{\hh}{\mathcal H}
\newcommand{\kk}{\mathcal K}

\newcommand{\la}{\lambda}
\newcommand{\eps}{\epsilon}
\newcommand{\lspan}[1]{\operatorname{span}\{#1\}}

\newcommand{\argmin}[1]{\mathop{\operatorname{argmin}}_{#1}}

\newcommand{\X}{{X}}

\newcommand{\EE}{{\mathcal E}}

\newcommand{\rhox}{{\rho_{\X}}}

\renewcommand{\C}{{C}}
\renewcommand{\k}{{K}}
\newcommand{\K}{K_n}
\newcommand{\Cn}{{\C}_n}
\newcommand{\Cl}{\C_\la}
\newcommand{\Cnl}{{\C}_{n\lambda}}
\newcommand{\Cm}{{\C}_m}
\newcommand{\Sn}{S_n}
\newcommand{\Zm}{Z_m}
\newcommand{\Pm}{P_m}
\newcommand{\yn}{\widehat y_n}

\newcommand{\eqals}[1]{{\begin{align*}#1\end{align*}}}
\newcommand{\eqal}[1]{{\begin{align}#1\end{align}}}
\newcommand{\bpr}{\begin{proof}}
\newcommand{\epr}{\end{proof}}
\newcommand{\be}{\begin{equation}}
\newcommand{\ee}{\end{equation}}

\newtheorem{definition}{Definition}
\newcommand{\bd}{\begin{definition}}
\newcommand{\ed}{\end{definition}}

\newcommand{\bi}{\begin{itemize}}
\newcommand{\ei}{\end{itemize}}

\newtheorem{ass}{Assumption}
\newcommand{\ba}{\begin{ass}}
\newcommand{\ea}{\end{ass}}

\newtheorem{remark}{Remark}
\newcommand{\br}{\begin{remark}}
\newcommand{\er}{\end{remark}}

\newtheorem{proposition}{Proposition}
\newcommand{\bp}{\begin{proposition}}
\newcommand{\ep}{\end{proposition}}

\newtheorem{lemma}{Lemma}
\newcommand{\blm}{\begin{lemma}}
\newcommand{\elm}{\end{lemma}}

\newtheorem{theorem}{Theorem}
\newcommand{\bt}{\begin{theorem}}
\newcommand{\et}{\end{theorem}}

\newtheorem{corollary}{Corollary}
\newcommand{\bcor}{\begin{corollary}}
\newcommand{\ecor}{\end{corollary}}

\begin{document}

\maketitle

\begin{bibunit}[unsrt]
\begin{abstract}
We study Nystr\"om type subsampling approaches  to large   scale  kernel methods, and  prove   learning bounds in the  statistical learning setting,  where random sampling and high probability estimates are considered.   In particular, we prove that these approaches  can achieve optimal learning bounds, provided the subsampling level is suitably chosen. These results suggest a simple  incremental variant of Nystr\"om Kernel Regularized Least Squares, where the subsampling level implements a form of computational regularization, in the sense that it controls at the same time  regularization and computations.  Extensive experimental analysis shows that the considered approach achieves state of the art performances on benchmark large scale datasets. 
\end{abstract}

\section{Introduction}

Kernel methods provide an elegant and effective framework to develop nonparametric 
statistical approaches to learning \cite{schlkopf2002learning}. However, memory requirements make these methods unfeasible when dealing with large datasets. Indeed, this observation has motivated a variety of  computational strategies to develop large scale kernel methods \cite{conf/icml/SmolaS00,conf/nips/WilliamsS00,conf/nips/RahimiR07,conf/icml/YangSAM14,conf/icml/LeSS13,conf/icml/SiHD14,conf/colt/ZhangDW13}. \\
In this paper we study subsampling methods, that we broadly refer to as Nystr\"om 
approaches. These methods replace the empirical kernel matrix, 
needed by standard kernel methods, with a smaller matrix obtained by (column) subsampling \cite{conf/icml/SmolaS00,conf/nips/WilliamsS00}. Such  procedures are shown to often dramatically reduce memory/time requirements while preserving good practical performances \cite{conf/nips/KumarMT09,conf/icml/LiKL10,Zhang:2008:INL:1390156.1390311,conf/nips/DaiXHLRBS14}.
The goal of our study is two-fold. First, and foremost,  we aim at providing a theoretical characterization of  the generalization properties of such  learning schemes
in a statistical learning setting. Second,  we wish to understand the
role played by the subsampling level both from a statistical and a computational point of view. As discussed in the following, this latter question leads to a natural variant of Kernel Regularized Least Squares (KRLS), where the subsampling level controls both regularization and computations. 

From a theoretical perspective, the effect of  Nystr\"om approaches has been primarily characterized considering the discrepancy between a given empirical kernel matrix and its subsampled version \cite{Drineas:2005:NMA:1046920.1194916,gittens2013revisiting,Wang:2013:ICM:2567709.2567748,journals/jmlr/DrineasMMW12,conf/innovations/CohenLMMPS15,conf/aistats/WangZ14,Kumar:2012:SMN:2503308.2343678}. While interesting in their own right, these latter results do not directly  yield information on the generalization properties of the obtained algorithm. Results in this direction, albeit suboptimal,  were first derived in \cite{journals/jmlr/CortesMT10} (see also \cite{6547995,conf/nips/YangLMJZ12}), and more recently in \cite{conf/colt/Bach13,alaoui2014fast}. 
In these latter papers, sharp error analyses in expectation are derived  in a fixed design  regression setting for a form of Kernel Regularized Least Squares. In particular, in  \cite{conf/colt/Bach13} a basic uniform sampling approach is studied, while in \cite{alaoui2014fast} a subsampling scheme based on the notion of leverage score is considered.   The main technical contribution of our study is an extension of these latter results to the statistical learning setting, where  the design is random and  high probability estimates are considered. The more general setting makes the analysis considerably more complex. Our main result gives optimal finite sample bounds for both uniform and leverage score based subsampling strategies. These methods are shown to achieve the same (optimal) learning error as kernel regularized least squares, recovered as a special case, while allowing substantial computational gains. Our analysis highlights the interplay  between the regularization and subsampling parameters, suggesting 
that the latter can be used to control simultaneously  regularization and computations. This strategy implements a form of {\em computational regularization} in the sense that the computational resources are tailored to the generalization properties in the data. This idea is developed  considering an incremental strategy to efficiently compute   learning solutions for different subsampling levels. The  procedure thus obtained,  which is a simple variant of classical Nystr\"om Kernel Regularized Least Squares with uniform sampling,  allows for efficient  model selection and achieves state of the art results on a variety of benchmark large scale datasets. \\
The rest of the paper is organized as follows. In Section \ref{sect:krls-nyst}, we introduce the setting and algorithms we consider.
In Section \ref{sect:theo-analysis}, we present our main theoretical contributions. In Section \ref{sect:incAlgoExperiments}, we discuss computational aspects and experimental results.
%
%
%
%
%
\section{Supervised learning with KRLS and  \Nystrom{} approaches} \label{sect:krls-nyst}
Let $\X\times \R$ be a probability space with distribution $\rho$, where we view $\X$ 
and $\R$ as the input and output spaces, respectively. Let $\rhox$ denote the marginal distribution of $\rho$ on $\X$ and $\rho(\cdot|x)$ the conditional distribution on $\R$ given $x\in \X$. Given a hypothesis space $\hh$ of measurable functions from $\X$ to $\R$, the goal is to minimize the {\em expected risk},
\be\label{eq:expmin}
\min_{f\in \hh} \EE(f), \quad\quad  \EE(f) = \int_{\X\times \R} (f(x)-y)^2 d\rho(x,y),
\ee
provided $\rho$ is known only through a training set of $(x_i, y_i)_{i=1}^n$ sampled identically and independently according to $\rho$.  A basic example of the above setting is random design regression with the squared loss, in which case
\be\label{eq:rdreg}
y_i=f_*(x_i)+\eps_i, \quad i=1, \dots, n,
\ee
with $f_*$ a fixed {\em regression} function, $\eps_1, \dots, \eps_n$ a sequence of  random variables seen as noise, and $x_1, \dots, x_n$ random inputs.
%
In the following, we consider kernel methods,  based on choosing a hypothesis space which is a separable reproducing kernel Hilbert space. The latter is a Hilbert space $\hh$ of functions, with inner product $\scal{\cdot}{\cdot}_\hh$, such that there exists a function $K:\X \times \X\to \R$ with the following two properties: 1) for all $x\in \X$, $K_x(\cdot)=K(x, \cdot)$ belongs to $\hh$,  and 2) the so called reproducing property holds: $f(x)=\scal{f}{K_x}_\hh$, for all $f\in \hh,\, x\in \X$ \cite{steinwart2008support}.
The function $K$, called reproducing kernel,  is easily shown to be symmetric and positive definite, that is the kernel matrix $(K_N)_{i,j}=K(x_i,x_j) $ is positive semidefinite for all $x_1, \dots, x_N\in \X$, $N \in \N$.
A classical way to derive an empirical solution to problem~\eqref{eq:expmin} is to consider 
a Tikhonov regularization approach,  based on the   minimization of the penalized empirical functional,
\be\label{eq:krls-problem}
\min_{f \in \hh} \frac{1}{n}\sum_{i=1}^{n}(f(x_i)-y_i)^2 + \la \nor{f}^2_\hh, \la > 0.
\ee
The above approach is referred to as Kernel Regularized Least Squares (KRLS) or Kernel Ridge Regression (KRR).  It is easy to see that a solution $\hat{f}_\la$ to problem~\eqref{eq:krls-problem} exists, it is unique and the representer theorem \cite{schlkopf2002learning} shows that it can be written as 
 \be\label{eq:rep}
\hat{f}_\la(x) = \sum_{i=1}^n \hat\alpha_i K(x_i, x) \quad \textrm{ with } \quad \hat\alpha = ({\K} + \la n I)^{-1} y, 
\ee
where $x_1,\dots, x_n$ are the training set points, $y = (y_1,\dots,y_n)$ and ${\K}$ is the empirical kernel matrix. Note that this result implies that we can restrict the minimization in~\eqref{eq:krls-problem} to the space,
$$\hh_n = \{f \in \hh~|~ f = \sum_{i=1}^n \alpha_i K({x}_i, \cdot),\; \alpha_1, \dots, \alpha_n \in \R\}.$$
Storing the kernel matrix ${\K}$, and solving the linear system in~\eqref{eq:rep},  can become computationally unfeasible as  $n$ increases. In the following, we consider strategies to find more efficient solutions, based on the idea of replacing $\hh_n$ with 
$$\hh_m = \{f ~|~ f = \sum_{i=1}^m \alpha_i K(\tilde{x}_i, \cdot),\; \alpha \in \R^m\},$$
where $m\le n$ and $\{\tilde{x}_1, \dots, \tilde{x}_m\}$ is a subset of the input  points in the training set. The solution $\hat{f}_{\la, m}$ of the corresponding minimization problem can now be written as, 
\be \label{eq:repny}
\hat{f}_{\la, m}(x) = \sum_{i=1}^m \tilde{\alpha}_i K(\tilde{x}_i, x)\quad \textrm{with}\quad
\tilde{\alpha} = (K_{nm}^\top K_{nm} + \la n K_{mm})^\dag K_{nm}^\top y,
\ee 
where $A^\dag$ denotes the Moore-Penrose pseudoinverse of a matrix $A$, and $(K_{nm})_{ij} = K(x_i, \tilde{x}_j)$, $(K_{mm})_{kj} = K(\tilde{x}_k, \tilde{x}_j)$ with $i \in \{1,\dots,n\}$ and $j,k \in \{1,\dots,m\}$  \cite{conf/icml/SmolaS00}. The above approach is related to \Nystrom{} methods and   different approximation strategies  correspond to different ways to select the inputs subset. While our framework applies to a broader class of strategies, see Section~\ref{sec:ext}, in the following we primarily consider two techniques.\\
{\bf Plain \Nystrom{}}. The points $\{\tilde{x}_1, \dots, \tilde{x}_m\}$ are sampled uniformly at random without replacement from the training set.\\
{\bf Approximate leverage scores (ALS) \Nystrom{}}.
Recall that the {\em leverage scores} associated to the training set points $x_1, \dots, x_n$  are 
 \be \label{eq:levscoredef}
 (l_i(t))_{i=1}^n, \quad l_i(t) = (\K (\K + t n I)^{-1})_{ii}, \quad i \in \{1, \dots, n\}
 \ee 
for any $t > 0$, where $(\K)_{ij} = K(x_i, x_j)$. In practice, leverage scores are onerous to compute and approximations
$(\hat l_i(t))_{i=1}^n$ can be considered \cite{journals/jmlr/DrineasMMW12,alaoui2014fast,conf/innovations/CohenLMMPS15} . In particular, in the following we are interested in suitable approximations defined as follows:
\bd[$T$-approximate leverage scores]\label{def:approx-lev-scores}
Let $(l_i(t))_{i=1}^n$ be the leverage scores associated to the training set for a given $t$. Let $\delta > 0$, $t_0 > 0$ and $T \geq 1$. We say that $(\widehat l_i(t))_{i=1}^n$ are $T$-approximate leverage scores with confidence $\delta$, when with probability at least $1-\delta$, 
$$
\frac{1}{T} l_i(t) \leq \widehat l_i(t) \leq T l_i(t) \quad \forall i\in\{1,\dots,n\}, t \geq t_0.
$$
\ed
Given $T$-approximate leverage scores for $t > \la_0$,  $\{\tilde{x}_1, \dots, \tilde{x}_m\}$  are sampled from the training set  independently with replacement, and  with  probability to be selected given by  $P_t(i) =  \hat l_i(t) / \sum_j \hat l_j(t)$. 
\\
In the next section, we state and discuss our main result showing that the KRLS formulation based on  plain or approximate leverage scores \Nystrom{} provides optimal empirical solutions to problem~\eqref{eq:expmin}.
%
\section{Theoretical analysis}\label{sect:theo-analysis}
In this section, we state and discuss our main results. We need several assumptions.
The first  basic assumption is that problem~\eqref{eq:expmin} admits at least a  solution.
\ba\label{as:exists-fh} 
There exists an $f_\hh \in \hh$ such that 
$$\EE(f_\hh)=\min_{f\in \hh}\EE (f).$$
\ea
Note that,  while the minimizer might not be unique, our results apply to the case in which $f_\hh$ is the unique minimizer with minimal norm. Also,  note that the above condition is weaker than assuming the regression function in~\eqref{eq:rdreg} to belong to $\hh$. Finally, we note that the study   of the paper can be adapted to  the case in which minimizers do not exist, but the analysis is considerably more involved and left to a longer version of the paper.
\\
The second assumption is a basic condition on the probability distribution.
\ba\label{as:noise}
Let $z_x$ be the random variable $z_x = y - f_\hh(x)$, with $x \in \X$, and $y$ distributed according to $\rho(y|x)$. Then, there exists $M, \sigma > 0$ such that $\mathbb{E} |z_x|^p \leq \frac{1}{2}p!M^{p-2}\sigma^2$ for any $p \geq 2$, almost everywhere on $\X$.
\ea
The above assumption is needed to control random quantities and is related to a {\em noise} assumption  in the regression model~\eqref{eq:rdreg}. It is clearly weaker than the often considered bounded output assumption \cite{steinwart2008support}, and trivially verified in classification.
\\
The last two assumptions describe the capacity (roughly speaking the {\em ``size''}) of the  hypothesis space induced by  $K$ with respect to  $\rho$ and the regularity of $f_\hh$ with respect to $K$ and $\rho$. 
To discuss them,  we first need the following definition.
\bd[Covariance operator and effective dimensions]
We define the covariance operator as 
$$
C: \hh \to \hh, \quad \scal{f}{C g}_\hh = \int_\X f(x)g(x)d\rhox(x) \;\;, \quad \forall \, f, g \in \hh.
$$
Moreover, for $\la>0$, we define the random variable ${\cal N}_x(\la) = \scal{K_x}{(C+\la I)^{-1} K_x}_\hh$ with $x\in \X$ distributed according to $\rhox$ and let 
$$
{\cal N}(\la) = \mathbb{E}\,{\cal N}_x(\la), \quad\quad {\cal N}_\infty(\la)=\sup_{x\in \X} {\cal N}_x(\la).
$$
\ed
We add several comments. Note that $C$ corresponds to the second moment operator, but we refer to it as the covariance operator with an abuse of terminology. Moreover, note that ${\cal N}(\la)  = \tr (C(C+\la I)^{-1})$ (see \cite{caponnetto2007optimal}). This latter quantity, called effective dimension or degrees of freedom,  can be seen as a measure of the 
capacity of the hypothesis space. The quantity ${\cal N}_\infty(\la)$ can be seen to provide a uniform bound on the leverage scores in Eq.~\eqref{eq:levscoredef}. Clearly, ${\cal N}(\la)\le {\cal N}_\infty(\la)$ for all $\la>0$.
\ba\label{as:kerrho}
The kernel $K$ is measurable, $C$ is bounded.
 Moreover, for all $\la>0$ and a $Q>0$,   
\eqal{
 & {\cal N}_\infty(\la)<\infty,\label{eq:lsbound}\\
 &{\cal N}(\la) \leq Q \la^{-\gamma}, \quad 0 < \gamma \leq 1.\label{eq:poleffdim}
}
\ea
Measurability of $K$ and boundedness of $C$  are  minimal conditions to ensure that the covariance operator is 
a well defined   linear, continuous, self-adjoint, positive operator \cite{steinwart2008support}. Condition~\eqref{eq:lsbound} is satisfied if the kernel is bounded
$\sup_{x\in \X}K(x,x)= \kappa^2<\infty$, indeed in this case $ {\cal N}_\infty(\la)\le \kappa^2/\la$ for all $\la>0$. Conversely, 
it can be seen that condition~\eqref{eq:lsbound} together with boundedness of $C$ imply  that the kernel is bounded, indeed
\footnote{If  ${\cal N}_\infty(\la)$ is finite, then ${\cal N}_\infty(\nor{C}) = \textrm{sup}_{x \in X} \nor{(C+\nor{C} I)^{-1}K_x}^2 \geq 1/2 \nor{C}^{-1}\textrm{sup}_{x \in X} \nor{K_x}^2$, therefore $K(x,x) \leq 2\nor{C}{\cal N}_\infty(\nor{C})$.}
$$
\kappa^2\le 2\nor{C}{\cal N}_\infty(\nor{C}).
$$
Boundedness of the kernel implies in particular that the operator $C$ is trace class  
and allows to  use tools from spectral theory. Condition~\eqref{eq:poleffdim} quantifies the capacity assumption and is related to 
covering/entropy number conditions (see \cite{steinwart2008support} for further details). In particular, it is known that condition~\eqref{eq:poleffdim}
is ensured if the eigenvalues $(\sigma_i)_i$ of $C$ satisfy a polynomial decaying condition $\sigma_i \sim i^{-\frac{1}{\gamma}}$.
Note that, since the operator $C$ is trace class, Condition~\eqref{eq:poleffdim} always holds for $\gamma=1$.
Here, for space constraints and in the interest of clarity we restrict  to such a polynomial  condition, but the analysis directly applies to other conditions including  exponential decay or a finite rank conditions \cite{caponnetto2007optimal}. 
Finally, we have the following regularity assumption. 
\ba\label{as:source}
 There exists $s \geq 0$, $1 \leq R < \infty$, such that $\nor{C^{-s} f_\hh}_{\hh} < R$. 
\ea
The above condition is fairly standard, and can be equivalently formulated in terms of classical concepts in approximation theory such as 
interpolation spaces \cite{steinwart2008support}. Intuitively, it quantifies the degree to which $f_\hh$ can be well approximated by functions in the RKHS $\hh$ and allows to control the bias/approximation error of a learning solution. For  $s=0$, it is always satisfied. For larger $s$,  we are assuming $f_\hh$ to belong to subspaces of $\hh$ that are the images of the  fractional compact operators $C^s$. Such spaces contain functions which, expanded on a basis of eigenfunctions of $C$, have larger coefficients in correspondence to large eigenvalues. Such an assumption is natural in view of using techniques such as~\eqref{eq:rep}, which  can be seen as a form  of spectral filtering, that estimate stable solutions by discarding the contribution of small eigenvalues \cite{journals/neco/GerfoROVV08}.  
In the next section, we are going to quantify the quality of empirical solutions of Problem~\eqref{eq:expmin} obtained by schemes of the form~\eqref{eq:repny}, in terms of the quantities in Assumptions~\ref{as:noise},~\ref{as:kerrho},~\ref{as:source}. 
%
\subsection{Main results}\label{sect:main-res}
In this section, we state and discuss our main results,  starting 
with optimal finite sample error bounds for regularized least squares based on plain  and approximate leverage score based \Nystrom{} subsampling.
\bt\label{thm:opt-rates-NyKRLS}
Under Assumptions~\ref{as:exists-fh}, \ref{as:noise}, ~\ref{as:kerrho},  and ~\ref{as:source}, 
let  $\delta>0$,  $v = \min(s, 1/2)$, $p = 1 + 1/(2v + \gamma)$ and assume
$$
n \,\geq\, 1655\kappa^2 + 223\kappa^2\log\frac{6\kappa^2}{\delta} + \left(\frac{38p}{\nor{C}} \log \frac{114\kappa^2 p}{\nor{C}\delta} \right)^p.$$ 
Then,  the following inequality  holds with probability at least $1-\delta$, 
\be\label{eq:excess-risk-bounded}
 \EE(\hat{f}_{\la, m}) - \EE(f_\hh) \leq q^2\, n^{-\frac{2v+1}{2v + \gamma + 1}}, \quad \textrm{with } q = 6R\left(2\nor{C}+\frac{M\kappa}{\sqrt{\nor{C}}} + \sqrt{\frac{Q\sigma^2 }{\nor{C}^\gamma}}\right)\log\frac{6}{\delta},
\ee
with  $\hat{f}_{\la, m}$  as in~\eqref{eq:repny}, $\la = \nor{C} n^{-\frac{1}{2v + \gamma + 1}}$ 
and 
\begin{enumerate}
\item for plain \Nystrom{} 
$$m \geq (67 \vee 5 {\cal N}_\infty(\la))\log\frac{12\kappa^2}{\la \delta};$$
\item for ALS \Nystrom{} and $T$-approximate leverage scores with subsampling probabilities $P_\la$, $t_0 \geq \frac{19\kappa^2}{n}\log\frac{12n}{\delta}$ and 
 $$m \geq  (334 \vee 78 T^2 {\cal N}(\la)) \log \frac{48n}{\delta}.$$
\end{enumerate}
\et
We add several comments.
First, the above results can be shown to be optimal in a minimax sense.
Indeed, minimax lower bounds proved in \cite{caponnetto2007optimal, SteinwartHS09} show that the learning rate in~\eqref{eq:excess-risk-bounded} is optimal under the considered assumptions (see Thm.~2,~3 of \cite{caponnetto2007optimal}, for a discussion on minimax lower bounds see Sec.~2 of \cite{caponnetto2007optimal}). Second, the obtained bounds can be compared to those obtained for other regularized learning techniques. Techniques known to achieve optimal error rates include Tikhonov regularization \cite{caponnetto2007optimal,SteinwartHS09,mendelson2010regularization}, iterative regularization by early stopping \cite{bauer,CapYao06}, spectral cut-off regularization (a.k.a. principal component regression or truncated SVD) \cite{bauer,CapYao06}, as well as regularized stochastic gradient methods \cite{yiming}.  All these techniques are essentially equivalent from a statistical point of view and differ only in the required computations. For example, iterative methods allow for a computation of solutions corresponding to different regularization levels which is more efficient than Tikhonov or SVD based approaches. 
The key observation is that  all these methods have the same $O(n^2)$ memory requirement. In this view, our results show that randomized subsampling methods can break such a memory barrier, and consequently achieve much better time complexity,  while preserving optimal learning guarantees. Finally,  we can compare our results with previous analysis of randomized kernel methods. As already mentioned, results close to those in Theorem~\ref{thm:opt-rates-NyKRLS} are given  in \cite{conf/colt/Bach13,alaoui2014fast} in a fixed design setting.  Our results extend and generalize the conclusions of these papers to a general statistical learning setting.  Relevant results are given in \cite{conf/colt/ZhangDW13} for a different  approach, based on averaging KRLS solutions obtained splitting the data in $m$ groups ({\em divide and conquer RLS}). The analysis in \cite{conf/colt/ZhangDW13} is only  in expectation, but considers random design and  
shows that the proposed method is indeed optimal provided the number of splits is chosen 
depending on the effective dimension $ {\cal N}(\la)$.
This is the only other work we are aware of establishing optimal learning rates for 
randomized kernel approaches in a statistical learning setting. In comparison with \Nystrom{} computational regularization the main disadvantage of the divide and conquer approach is computational and in the model selection phase where solutions corresponding to different regularization parameters and number of splits usually need to be computed. 
\\
The proof of Theorem~\ref{thm:opt-rates-NyKRLS} is fairly technical and lengthy. It incorporates ideas from \cite{caponnetto2007optimal} and techniques developed to study spectral filtering regularization \cite{bauer,rudi2013sample}. In the next section, we briefly sketch some main ideas and discuss how they suggest an interesting  perspective on 
regularization techniques including subsampling.
\begin{figure}[t]
{\centering
\subfigure{
                \includegraphics[trim=0cm 1cm 0cm 0.5cm, clip=true, width=0.3\textwidth]{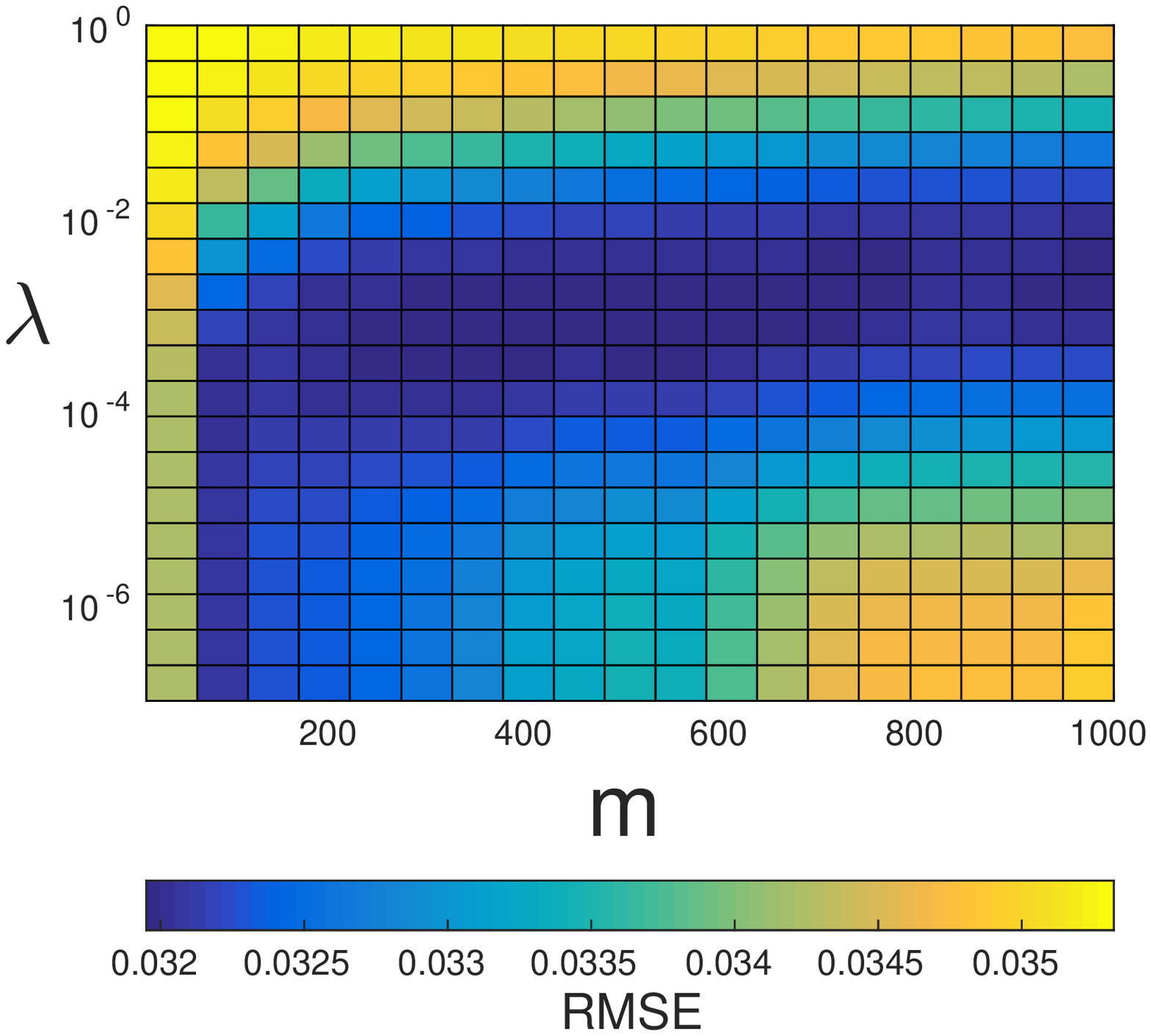}
        }
\subfigure{
                \includegraphics[trim=0cm 1cm 0cm 0.5cm, clip=true, width=0.3\textwidth]{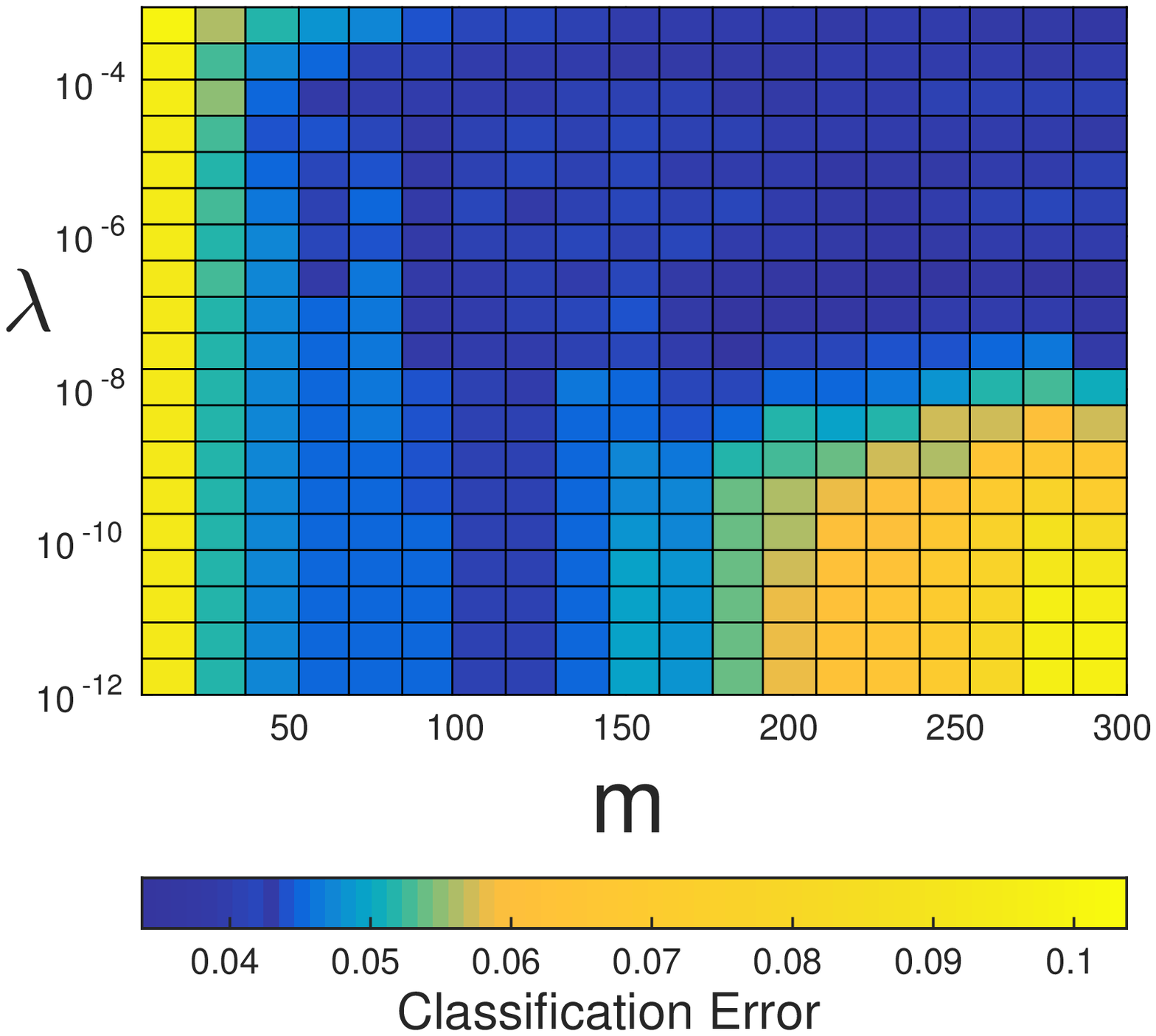}
        }
\subfigure{
                \includegraphics[trim=0cm 1cm 0cm 0.5cm, clip=true, width=0.3\textwidth]{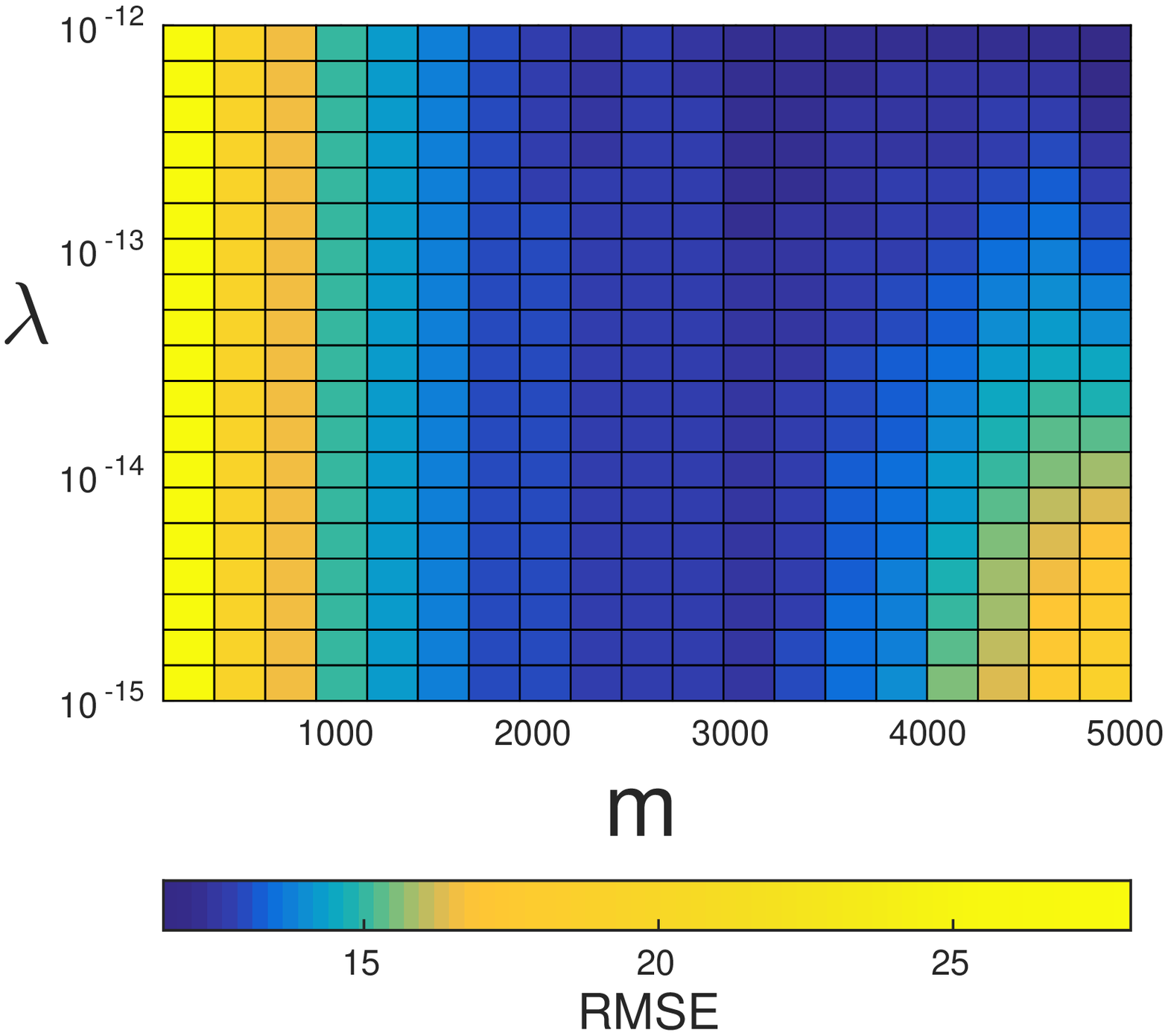}
        }
}
        \caption{Validation errors associated to $20 \times 20$ grids of values for $m$ (x axis) and $\lambda$ (y axis) on \texttt{pumadyn32nh} (left), \texttt{breast cancer} (center) and \texttt{cpuSmall} (right). \label{fig:m_lambda_plots}}
\end{figure}
\subsection{Proof sketch and a computational regularization perspective}
A key step in  the proof of Theorem~\ref{thm:opt-rates-NyKRLS} is an error decomposition, and corresponding bound, for any  fixed $\la$ and $m$. Indeed, it is proved in Theorem~\ref{thm:nys-err-dec} and Proposition~\ref{prop:bounds-plain-appr} that, for $\delta>0$, with probability at least $1-\delta$,
\be\label{eq:oracle}
\left|{\cal E}(\hat f_{\la , m}) - {\cal E}(f_{\hh})\right|^{1/2} \lesssim R\left(\frac{M \sqrt{{\cal N}_\infty(\la)}}{n} + \sqrt{\frac{\sigma^2 {\cal N}(\la)}{n}}\right)\log \frac{6}{\delta} +  R {\cal C}(m)^{1/2 + v} + R\la^{1/2+v}. 
\ee
The first and last term in the right hand side of the above inequality can be seen as forms of {\em sample and approximation errors} \cite{steinwart2008support} and are studied in Lemma~\ref{lm:sampling-error} and Theorem~\ref{thm:nys-err-dec}. 
The mid term can be seen as a {\em computational error} and depends on the considered subsampling scheme. 
Indeed, it is shown in Proposition~\ref{prop:bounds-plain-appr} that ${\cal C}(m)$ can be taken as, 
$$
{\cal C}_{\rm pl}(m) = \min \left\{t > 0 ~\middle|~ (67 \vee 5 {\cal N}_\infty(t))\log\frac{12\kappa^2}{t \delta} \leq m \right\},
$$
for the plain \Nystrom{} approach, and 
$$
{\cal C}_{\rm ALS}(m) = \min \left\{\frac{19\kappa^2}{n}\log\frac{12n}{\delta} \leq t \leq \nor{C}{} ~\middle|~ 78 T^2 {\cal N}(t) \log \frac{48n}{\delta} \leq m \right\},
$$
for the approximate leverage scores approach. The bounds in Theorem~\ref{thm:opt-rates-NyKRLS} follow by: 1) minimizing in $\la$ the sum of the first and third term 2) choosing $m$ so that the computational error is of the same order of the other terms. Computational resources and regularization are then tailored to the generalization properties of the data at hand.  We add a few comments. First, note that the error bound in~\eqref{eq:oracle} holds for a large class of subsampling schemes, as discussed in Section~\ref{sec:ext} in the appendix. Then specific error bounds can be derived developing computational error estimates. Second, the error bounds in 
Theorem~\ref{thm:nys-err-dec} and Proposition~\ref{prop:bounds-plain-appr}, and hence in Theorem~\ref{thm:opt-rates-NyKRLS}, easily generalize to a larger class of regularization schemes beyond Tikhonov approaches, namely spectral filtering \cite{bauer}. For space constraints, these  extensions are deferred to a longer version of the paper. Third, we note that, in practice, optimal data driven parameter choices, e.g. based on hold-out estimates \cite{CapYao06}, can be used to adaptively achieve optimal learning bounds.\\
Finally, we observe that a different perspective is derived starting from inequality~\eqref{eq:oracle}, and noting that the role played by $m$ and $\la$ can also be exchanged. Letting $m$ play the role of a regularization parameter, $\la$ can be set as a function of $m$ and $m$ tuned adaptively.  For example, in the case of a plain \Nystrom{} approach, if we set 
$$\la = \frac{\log m}{m},  \quad \text{and}\quad  m = 3 n^{\frac{1}{2v + \gamma + 1}}\log n,$$
then the obtained learning solution achieves the error bound in Eq.~\eqref{eq:excess-risk-bounded}. As above, the subsampling level can also  be chosen by cross-validation. Interestingly, in this case by tuning $m$ we naturally control computational resources and regularization.
An advantage of this latter parameterization is that, as described in the following, the solution corresponding 
to different subsampling levels is easy to update using Cholesky rank-one update formulas \cite{Golub1996}. As discussed in the next section, in practice, a joint tuning over $m$
and $\la$ can be done starting from small $m$ and appears to be advantageous both for error and computational performances. 
%
\section{Incremental updates and experimental analysis}\label{sect:incAlgoExperiments}
In this section, we first describe an incremental  strategy to efficiently  explore different subsampling levels
and then perform extensive empirical tests aimed in particular at: 1) investigating the statistical and computational benefits of considering varying subsampling levels, and 2) compare the performance of the algorithm with respect to state of the art solutions on several large scale benchmark datasets. Throughout this section, we only consider a plain \Nystrom{} approach, deferring to future work the
analysis of leverage scores based sampling techniques. Interestingly, we will see that such a basic approach can often provide state of the art performances.
%
\begin{figure}[t]
\begin{minipage}[b]{0.6\linewidth}
\begin{algorithmic}
 \State {{\bf Input:} Dataset $(x_i, y_i)_{i=1}^n$, Subsampling $(\tilde{x}_j)_{j=1}^m$,\\ Regularization Parameter $\la$.}
 \State  {{\bf Output:} \Nystrom{} KRLS estimators $\{\tilde\alpha_1,\dots,\tilde\alpha_m\}$.}
 \State Compute $\gamma_1$; $R_1 \gets \sqrt{\gamma_1};$
 \For{$t \in \{2,\dots, m\}$}
  \State Compute $A_t, u_t, v_t$;
  \State $R_t \gets \begin{pmatrix} R_{t-1} & 0\\ 0 & 0\end{pmatrix};\; 
  \begin{array}{l}R_t \gets {\tt cholup}(R_t, u_t, '+');\\
  R_t \gets {\tt cholup}(R_t, v_t, '-');\\
  \end{array}$
  \State $\tilde{\alpha}_t \gets R_t^{-1} (R_t^{-\top} (A_t^\top y));$
 \EndFor
\end{algorithmic}
\captionof{algorithm}{Incremental \Nystrom{} KRLS.\label{alg:incr-nys-krls}}
\end{minipage}
\hfill
\begin{minipage}[b]{0.39\linewidth}
\centering
\includegraphics[width=0.75\textwidth]{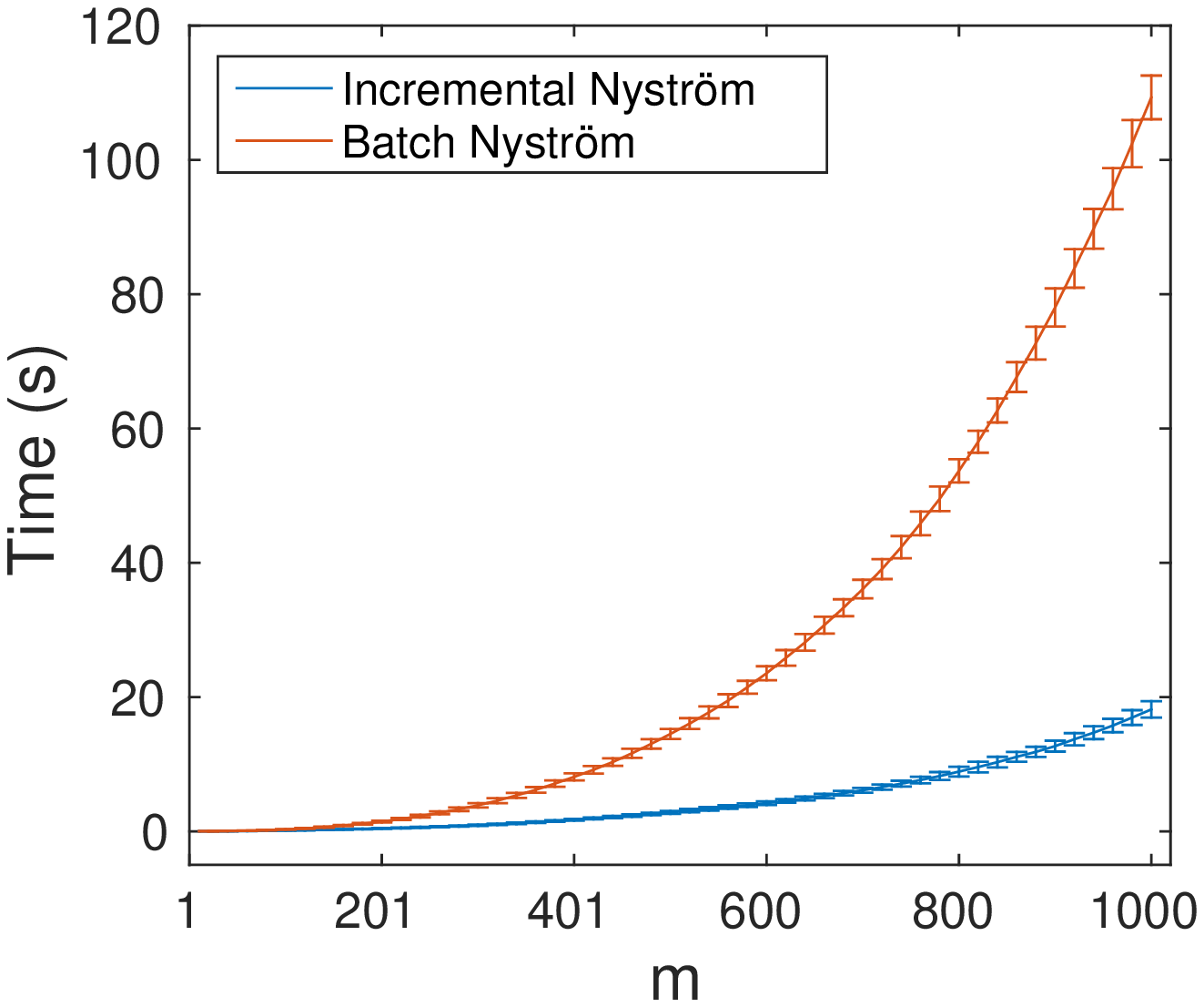}
\captionof{figure}{Model selection time on the \texttt{cpuSmall} dataset. $m \in \left[1,1000\right]$ and $T=50$, 10 repetitions.\label{fig:modSel1}}
\end{minipage}
\end{figure}
\subsection{Efficient incremental updates}\label{sect:eff-inc-updates}
Algorithm~\ref{alg:incr-nys-krls} efficiently computes 
solutions corresponding to different subsampling levels, by exploiting rank-one Cholesky updates \cite{Golub1996}. The proposed procedure allows to efficiently compute a whole regularization path of solutions, and hence perform fast model selection\footnote{The code for Algorithm~\ref{alg:incr-nys-krls} is available at \url{lcsl.github.io/NystromCoRe}.} (see Sect.~\ref{sect:alg}). In Algorithm~\ref{alg:incr-nys-krls}, the function {\tt cholup} is the Cholesky rank-one update formula available in many linear algebra libraries. The total cost of the algorithm is $O(nm^2 + m^3)$ time to compute $\tilde{\alpha}_2, \dots, \tilde{\alpha}_m$, while a naive non-incremental algorithm would require $O(nm^2T + m^3T)$ with $T$ is the number of analyzed subsampling levels. The following are some quantities needed by the algorithm: $A_1 = a_1$ and $A_t = (A_{t-1} \; a_t) \in \R^{n\times t}$, for any $2 \leq t \leq m$. Moreover, for any $1 \leq t \leq m$, $g_t = \sqrt{1 + \gamma_t}$ and
\eqals{
& u_t = (c_t/(1 + g_t),\,g_t),& & a_t = (K(\tilde{x}_t,x_1),\dots,K(\tilde{x}_t,x_n)),  & & c_t = A_{t-1}^\top a_t + \la n b_t, \\
& v_t = (c_t/(1 + g_t),\,-1),& & b_t = (K(\tilde{x}_t,\tilde{x}_1),\dots,K(\tilde{x}_t,\tilde{x}_{t-1})), & & \gamma_t = a_t^\top a_t + \la n K(\tilde{x}_t,\tilde{x}_t).
}
\subsection{Experimental analysis}
 We empirically study the properties of Algorithm~\ref{alg:incr-nys-krls}, considering a Gaussian kernel of width $\sigma$. The selected datasets are already divided in a training and a test part\footnote{In the following we denote by $n$ the total number of points and by $d$ the number of dimensions.}. We randomly split the training part in a training set and a validation set ($80\%$ and $20\%$ of the $n$ training points, respectively) for parameter tuning via cross-validation. The $m$ subsampled points for \Nystrom{} approximation are selected uniformly at random from the training set. We report the performance of the selected model on the fixed test set, repeating the process for several trials.\\
{\bf Interplay between $\la$ and $m$.} We begin with  a  set of results showing that incrementally  exploring different subsampling levels can yield 
very good performance while substantially reducing the computational requirements. We consider the
\texttt{pumadyn32nh} ($n=8192$, $d=32$), the  \texttt{breast cancer} ($n=569$, $d=30$), and   the \texttt{cpuSmall} ($n=8192$, $d=12$) datasets\footnote{\url{www.cs.toronto.edu/~delve} and \url{archive.ics.uci.edu/ml/datasets}}.  In Figure \ref{fig:m_lambda_plots}, we report the validation errors associated to a $20 \times 20$ grid of values for $\lambda$ and $m$. The $\lambda$ values are logarithmically spaced, while the $m$ values are linearly spaced. The ranges and kernel bandwidths, chosen according to preliminary tests on the data, are $\sigma = 2.66$, $\lambda \in \left[10^{-7}, 1\right]$, $m \in \left[ 10, 1000 \right]$ for \texttt{pumadyn32nh},  $\sigma = 0.9$, $\lambda \in \left[10^{-12},10^{-3}\right]$, $m \in \left[5, 300\right]$ for \texttt{breast cancer}, and $\sigma = 0.1$, $\lambda \in \left[10^{-15}, 10^{-12} \right]$, $m \in \left[100, 5000 \right]$ for \texttt{cpuSmall}. The main observation that can be derived from this first series of tests is that a small $m$ is sufficient to obtain the same results achieved with the largest $m$. For example, for \texttt{pumadyn32nh} it is sufficient to choose $m=62$ and $\lambda = 10^{-7}$ to obtain an average test RMSE of $0.33$ over 10 trials, which is the same as the one obtained using $m=1000$ and $\lambda = 10^{-3}$, with a 3-fold speedup of the joint training and validation phase. Also, it is interesting to observe that for given values of $\lambda$, large values of $m$ can decrease the performance. This observation is consistent with the results in Section~\ref{sect:main-res}, showing that $m$ can play the role of a regularization parameter.  Similar results are obtained for  \texttt{breast cancer}, where for $\lambda=4.28 \times 10^{-6}$ and $m=300$ we obtain a $1.24\%$ average classification error on the test set over 20 trials, while for $\lambda=10^{-12}$ and $m=67$ we obtain $1.86\%$. For \texttt{cpuSmall}, with $m=5000$ and $\lambda=10^{-12}$ the average test RMSE over 5 trials is $12.2$, while for $m=2679$ and $\lambda=10^{-15}$ it is only slightly higher, $13.3$, but computing its associated solution requires less than half of the time and approximately half of the memory.
%
\begin{table*}[t]
\caption{Test RMSE comparison for exact and approximated kernel methods. The results for KRLS, Batch \Nystrom{}, RF and Fastfood are the ones reported in \cite{conf/icml/LeSS13}. $n_{tr}$ is the size of the training set.\label{tab:exp1.1}}
\begin{center}
\resizebox{\textwidth}{!}{%
\begin{tabular}{ccccccccccc}
\toprule 
{\em Dataset} & $n_{tr}$ & $d$ & {\em Incremental} &{\em KRLS}   & {\em Batch } & {\em RF} & {\em Fastfood} & {\em Fastfood} & {\em KRLS} & {\em Fastfood} \\ 
&  &  & {\em \Nystrom{} RBF} & {\em RBF}  & {\em \Nystrom{} RBF} & {\em RBF} & {\em RBF} & {\em FFT} & {\em Matern} & {\em Matern}  \\ 
\midrule
Insurance Company & 5822 & 85 & $0.23180\pm 4 \times 10^{-5}$ & \textbf{0.231} & 0.232 & 0.266 & 0.264 & 0.266 & 0.234 & 0.235\\ 
CPU & 6554 & 21 & \textbf{$\mathbf{2.8466 \pm 0.0497}$} & 7.271 & 6.758 & 7.103 & 7.366 & 4.544 & 4.345 & 4.211\\  
CT slices (axial) & 42800 & 384 & $\mathbf{7.1106 \pm 0.0772}$ & NA & 60.683 & 49.491 & 43.858 & 58.425 & NA & 14.868\\ 
Year Prediction MSD & 463715 & 90 & $\mathbf{0.10470 \pm 5 \times 10^{-5}}$ & NA & 0.113 & 0.123 & 0.115 & 0.106 & NA & 0.116\\ 
Forest & 522910 & 54 & $0.9638 \pm 0.0186$ & NA & \textbf{0.837} & 0.840 & 0.840 & 0.838 & NA & 0.976\\ 
\bottomrule
\hline 
\end{tabular} 
}
\end{center}
\end{table*}
\\{\bf Regularization path computation}. If the subsampling level $m$ is used as a regularization parameter, the computation of a regularization path corresponding to different subsampling levels becomes crucial during the model selection phase. A naive approach, that consists in recomputing the solutions of Eq.~\ref{eq:repny} for each subsampling level, 
 would require $O(m^2nT + m^3LT)$ computational time, where $T$ is the number of solutions with different subsampling levels to be evaluated and $L$ is the number of Tikhonov regularization parameters.
  On the other hand, by using the incremental \Nystrom{} algorithm the model selection time complexity is $O(m^2n + m^3L)$ for the whole regularization path. We experimentally verify this speedup on \texttt{cpuSmall} with 10 repetitions, setting $m \in \left[1,5000\right]$ and $T=50$. The model selection times, measured on a server with 12 $\times$ 2.10GHz Intel$^\circledR$ Xeon$^\circledR$ E5-2620 v2  CPUs and 132 GB of RAM, are reported in Figure \ref{fig:modSel1}. The result clearly confirms the beneficial effects of incremental \Nystrom{} model selection on the computational time.\\
{\bf Predictive performance comparison}.
Finally,  we consider  the performance of the algorithm on several large scale benchmark datasets considered in \cite{conf/icml/LeSS13}, see Table \ref{tab:exp1.1}. $\sigma$ has been chosen on the basis of preliminary data analysis. $m$ and $\la$ have been chosen by cross-validation, starting from small subsampling values up to $m_{max}=2048$, and considering $\la \in \left[ 10^{-12} , 1\right]$. After model selection, we retrain the best model on the entire training set and compute the RMSE on the test set. We consider 10 trials, reporting the performance mean and standard deviation. The results in Table \ref{tab:exp1.1} compare \Nystrom{} computational regularization with the following methods (as in \cite{conf/icml/LeSS13}):
\begin{itemize}
\item \textbf{Kernel Regularized Least Squares (KRLS):} Not compatible with large datasets.
\item \textbf{Random Fourier features (RF):} As in \cite{conf/nips/RahimiR07}, with a number of random features $D=2048$.
\item \textbf{Fastfood RBF, FFT and Matern kernel:} As in \cite{conf/icml/LeSS13}, with $D=2048$ random features.
\item \textbf{Batch \Nystrom{}:} \Nystrom{} method \cite{conf/nips/WilliamsS00} with uniform sampling and $m=2048$.
\end{itemize}
The above results show that the proposed incremental \Nystrom{} approach behaves really well, matching state of the art predictive performances.
%
\subsubsection*{Acknowledgments}
The work described in this paper is  supported by the Center for Brains, Minds and Machines (CBMM), funded by NSF STC award CCF-1231216; and by FIRB project RBFR12M3AC, funded by the Italian Ministry of Education, University and Research. 
{\small
\putbib[nipsBib]
}
\end{bibunit}
\newpage
\appendix
\begin{bibunit}[unsrt]


\section{The incremental algorithm}\label{sect:alg}
Let $(x_i, y_i)_{i=1}^{n}$ be the dataset and $(\tilde{x}_i)_{i=1}^{m}$ be the selected \Nystrom{} points. 
We want to compute $\tilde \alpha$ of Eq.~\ref{eq:repny}, incrementally in $m$. Towards this goal we compute an incremental Cholesky decomposition $R_t$ for $t \in \{1,\dots,m\}$ of the matrix $G_t = K_{n t}^\top K_{n t} + \la n K_{t t}$, and the coefficients $\tilde{\alpha}_t$ by $\tilde{\alpha}_t = R_t^{-1} R_t^{-\top}K_{n t}^\top y$. Note that, for any $1 \leq t \leq m-1$, by assuming $G_t = R_t^\top R_t$ for an upper triangular matrix $R_t$, we have
\eqals{
& G_{t+1} = \begin{pmatrix} G_t & c_{t+1}\\ c_{t+1}^\top & \gamma_{t+1}\end{pmatrix} = \begin{pmatrix} R_t & 0\\ 0 & 0\end{pmatrix}^\top \begin{pmatrix} R_t & 0\\ 0 & 0\end{pmatrix} + C_{t+1} \quad {\rm with} \quad C_{t+1} = \begin{pmatrix} 0 & c_{t+1}\\ c_{t+1}^\top & \gamma_{t+1}\end{pmatrix}, 
}
and $c_{t+1}$, $\gamma_{t+1}$ as in Section~\ref{sect:eff-inc-updates}.
Note moreover that $G_1 = \gamma_1$. Thus if we decompose the matrix $C_{t+1}$ in the form $C_{t+1} = u_{t+1} u_{t+1}^\top - v_{t+1} v_{t+1}^\top$ we are able compute $R_{t+1}$, the Cholesky matrix of $G_{t+1}$, by updating a bordered version of $R_t$ with two rank-one Cholesky updates. This is exactly Algorithm~\ref{alg:incr-nys-krls} with $u_{t+1}$ and $v_{t+1}$ as in Section~\ref{sect:eff-inc-updates}.
Note that the rank-one Cholesky update requires $O(t^2)$ at each call, while the computation of $c_t$ requires $O(n t)$ and the ones of $\tilde\alpha_t$ requires to solve two triangular linear systems, that is $O(t^2 + nt)$. Therefore the total cost for computing $\tilde\alpha_2,\dots,\tilde{\alpha}_m$ is $O(nm^2 + m^3)$.

\section{Preliminary definitions}\label{sect:op-def}
%
We begin introducing several operators that will be useful in the following. 
Let  $z_1,\dots, z_m \in \hh$ and  for all $f \in \hh$, $a \in \R^m$, let  
\begin{eqnarray*}
\Zm: \hh \to \R^m,& \quad \Zm f = (\scal{z_1}{f}_\hh,\dots,\scal{z_m}{f}_\hh), \\
\Zm^*:  \R^m \to\hh,& \quad \Zm^* a = \sum_{i=1}^m a_i z_i. 
\end{eqnarray*}

Let  $\Sn=\frac{1}{\sqrt{n}}\Zm$ and $\Sn^* = \frac{1}{\sqrt{n}}Z_m^*$  the operators obtained taking $m=n$ and $z_i=K_{x_i}$,  $\forall i=1, \dots, n$ in the above definitions.
Moreover,  for all $ f,g \in \hh$ let 
$$
\Cn: \hh \to \hh, \quad \scal{f}{C_n g}_\hh = \frac{1}{n} \sum_{i=1}^n f(x_i)g(x_i).
$$
The above operators are   linear and  finite rank. Moreover
$C_n = S^*_{n}S_{n}$ and $\K = n S_{n}S_{n}^*$, and further $B_{nm} = \sqrt{n} S_n \Zm^* \in \R^{n\times m}$, $G_{mm} = \Zm \Zm^* \in \R^{m\times m}$ and $\tilde K_n = B_{nm} G_{mm}^\dag B_{nm}^\top \in \R^{n\times n}$.

%
\section{Representer theorem for \Nystrom{} computational regularization and extensions}\label{sect:equiv-form}

In this section we consider explicit representations of the estimator obtained via 
\Nystrom{} computational regularization and extensions. Indeed, we consider 
 a general
subspace $\hh_m$ of $\hh$, and the following problem
\eqal{\label{eq:gen-ny-krls-problem-0}
\hat f_{\la, m} = \argmin{f \in \hh_m} \;\frac{1}{n}\sum_{i=1}^n(f(x_i) - y_i)^2 + \la \nor{f}^2_\hh.
}
In the following lemmas, we show three different characterizations of $f_{\la, m}$.
\blm\label{lm:char0-flam}
Let $f_{\la, m}$ be the solution of the problem in Eq.~\eqref{eq:gen-ny-krls-problem-0}. Then
it is characterized by the following equation
\eqal{\label{eq:general-eq-flam}
(\Pm \Cn \Pm + \la I) \hat f_{\la, m} = \Pm \Sn^* \yn,
}
with $\Pm$ the projection operator with range  $\hh_m$ and  $\yn = \frac{1}{\sqrt{n}} y$.
\elm
\bpr
The proof proceeds in three steps. First, note that, by rewriting  Problem~\eqref{eq:gen-ny-krls-problem-0} with the notation introduced in the previous section, we obtain,
\eqal{\label{eq:gen-ny-krls-problem}
\hat f_{\la, m} = \argmin{f \in \hh_m} \;\nor{S_n f - \yn}^2 + \la \nor{f}^2_\hh.
}
This problem is strictly convex and coercive, therefore admits a unique solution. Second, we  show that its solution coincide to the one of the following problem,
\eqal{\label{eq:gen-ny-krls-problem-1}
\hat f^* = \argmin{f \in \hh} \, \nor{S_n \Pm f - \yn}^2 + \la \nor{f}^2_\hh.
}
Note that the above problem is again strictly convex and coercive.
To show that $\hat f_{\la, m} = \hat f^*$, let $\hat f^* = a + b$ with $a \in \hh_m$ and $b \in \hh_m^\bot$. A necessary condition for $\hat f^*$ to be optimal, is that $b = 0$, indeed, considering that $\Pm b = 0$, we have
$$ \nor{S_n \Pm f^* - \yn}^2 + \la \nor{f^*}^2_\hh = \nor{S_n \Pm a - \yn}^2 + \la \nor{a}^2_\hh + \la \nor{b}^2_\hh \geq \nor{S_n \Pm a - \yn}^2 + \la \nor{a}^2_\hh.$$
This means that $\hat f^* \in \hh_m$, but on $\hh_m$ the functionals defining  Problem~\eqref{eq:gen-ny-krls-problem} and Problem~\eqref{eq:gen-ny-krls-problem-1} are identical because $\Pm f = f$ for any $f \in \hh_m$ and so $\hat f_{\la, m} = \hat f^*$. Therefore, by computing the derivative of the functional of Problem~\eqref{eq:gen-ny-krls-problem-1}, we see that $\hat f_{\la, m}$ is given  by Eq.~\eqref{eq:general-eq-flam}.
\epr
Using the above results, we can give an equivalent representations of the function $\hat{f}_{\la,m}$.
Towards this end, let $Z_m$ be a linear operator as in Sect.~\ref{sect:op-def} such that the range of $Z_m^*$ is exactly $\hh_m$. Morever, let
$$
\Zm = U\Sigma V^*
$$ 
be the SVD of $\Zm$ where $U: \R^t \to \R^m$, $\Sigma: \R^t \to \R^t$, $V:\R^t \to \hh$, $t \leq m$ and $\Sigma = \textrm{diag}(\sigma_1,\dots,\sigma_t)$ with $\sigma_1 \geq \dots \geq \sigma_t > 0$, $U^*U = I_t$ and $V^*V = I_t$. Then  the orthogonal projection operator $\Pm$ is given by  $\Pm = VV^*$ and  the range of $\Pm$ is exactly $\hh_m$.
In the following lemma we give a characterization of $\hat{f}_{\la,m}$ that will be useful in the proof of the main theorem.
\blm
\label{lm:char1-flam}
Given the above definitions , $\hat{f}_{\la,m}$ can be written as
\eqal{\label{eq:char1-flam}
\hat{f}_{\la,m} = V (V^* C_n V + \la I)^{-1} V^* \Sn^* \yn.
}
\elm
\bpr By Lemma~\ref{lm:char0-flam}, we know that $\hat f_{\la, m}$ is written as in Eq.~\eqref{eq:general-eq-flam}.
Now, note that $\hat f_{\la, m} = \Pm \hat f_{\la, m}$ and Eq.~\eqref{eq:general-eq-flam} imply $(\Pm \Cm \Pm + \la I) \Pm \hat f_{\la, m} = \Pm \Sn^* \yn$, 
that is equivalent to
$$V(V^* \Cn V + \la I)V^* \hat f_{\la, m} = VV^* \Sn^* \yn,$$
by substituting $\Pm$ with $VV^*$.
Thus by premultiplying the previous equation by $V^*$ and dividing by $V^* \Cm V + \la I$, we have
$$V^* \hat f_{\la, m} = (V^* \Cm V + \la I)^{-1}V^* \Sn^* \yn.$$
Finally, by premultiplying by $V$, 
$$\hat f_{\la, m} = \Pm \hat f_{\la, m} = V(V^* \Cm V + \la I)^{-1}V^* \Sn^* \yn.$$
\epr
Finally, the following result provide a characterization of the solution useful for computations.
\blm[Representer theorem for $\hat{f}_{\la,m}$]\label{lm:flm-rewriting}
Given the above definitions, we have that $\hat{f}_{\la,m}$ can be written as 
\eqal{\label{eq:char2-flam}
\hat{f}_{\la, m}(x) = \sum_{i=1}^m \tilde{\alpha}_i z_i(x), \quad \textrm{with }
\tilde{\alpha} = (B_{nm}^\top B_{nm} + \la n G_{mm})^\dag B_{nm}^\top y \quad\;\;\; \forall \; x \in X.
}
\elm
\bpr
According to the definitions of $B_{nm}$ and $G_{mm}$ we have that
$$\tilde\alpha = (B_{nm}^\top B_{nm} + \lambda n G_{mm})^\dag B_{nm}^\top y = ((\Zm \Sn^*)(\Sn\Zm^*) + \la (\Zm\Zm^*))^\dag (\Zm \Sn^*) {\widehat y_n}.$$
Moreover, according to the definition of $\Zm$ we have $$\hat f_{\la, m}(x) = \sum_{i=1}^m \tilde\alpha_i \scal{z_i}{K_{x}} =  \langle\Zm \k_x, \tilde\alpha\rangle_{\R^m} = \langle\k_x,\Zm^* \tilde\alpha\rangle_\hh \quad \forall \; x \in X,$$ so that 
$$
\hat{f}_{\la,m} = \Zm^* ((\Zm \Sn^*)(\Sn\Zm^*) + \la (\Zm\Zm^*))^\dag (\Zm \Sn^*) {\widehat y_n} = \Zm^* (\Zm \Cnl\Zm^*)^\dag (\Zm \Sn^*) {\widehat y_n},
$$
where $\Cnl = C_n + \la I$. Let $F = U\Sigma$, $G = V^*\Cn V + \la I$,  $H = \Sigma U^\top$, and note that $F$, $GH$, $G$ and $H$ are full-rank matrices, then we can perform the full-rank factorization of the pseudo-inverse (see Eq.24, Thm. 5, Chap. 1 of \cite{ben2003generalized}) obtaining
$$(\Zm \Cnl\Zm^*)^\dag  = (F G H)^\dag =H^\dag (FG)^\dag =  H^\dag G^{-1} F^\dag = U\Sigma^{-1} (V^* \Cn V + \la I)^{-1} \Sigma^{-1} U^*.$$
Finally,  simplyfing $U$ and $\Sigma$, we have
\eqals{
\hat{f}_{\la,m} & = \Zm^* (\Zm \Cnl\Zm^*)^\dag (\Zm \Sn^*) {\widehat y_n} \\
& = V \Sigma U^* U \Sigma^{-1} (V^* C_n V + \la I)^{-1} \Sigma^{-1} U^* U \Sigma V^* \Sn^* {\widehat y_n}\\
& = V (V^* C_n V + \la I)^{-1} V^* \Sn^* {\widehat y_n}.
}
\epr
\subsection{Extensions}\label{sec:ext}
Inspection of the proof shows that our analysis extends beyond the class of subsampling schemes in Theorem~\ref{thm:opt-rates-NyKRLS}. Indeed, the error decomposition Theorem~\ref{thm:nys-err-dec} directly applies to a large family of approximation schemes. Several further examples are described next.

\paragraph{KRLS and Generalized \Nystrom{}}
In general we could choose an arbitrary $\hh_m \subseteq \hh$. Let $\Zm: \hh \to  \R^m$ be a linear operator such that 
\be
\hh_m = \ran{\Zm^*} = \{f~|~ f = \Zm^*\alpha,\, \alpha \in \R^m\}.
\ee
Without loss of generality, $\Zm^*$ is expressible as $\Zm^* = (z_1,\dots,z_m)^\top$ with $z_1,\dots,z_m \in \hh$, therefore, according to Section~\ref{sect:op-def} and to Lemma~\ref{lm:flm-rewriting},
the solution of KRLS approximated with the generalized \Nystrom{} scheme is
\be\label{eq:def-gen-nystrom}
\hat{f}_{\la, m}(x) = \sum_{i=1}^m \tilde{\alpha}_i z_i(x), \quad \textrm{with }
\tilde{\alpha} = (B_{nm}^\top B_{nm} + \la n G_{mm})^\dag B_{nm}^\top y
\ee
with $B_{nm} \in \R^{n\times m}$, $(B_{nm})_{ij} = z_j(x_i)$ and $G_{mm} \in \R^{m \times m}$, $(G_{mm})_{ij} = \scal{z_i}{z_j}_\hh$, or equivalently
\be
\hat{f}_{\la, m}(x) = \sum_{i=1}^m \tilde{\alpha}_i z_i(x), \quad
\tilde{\alpha} =  G_{mm}^\dag B_{nm}^\top (\tilde K_n  + \la n I)^\dag \yn,\quad \tilde K_n = B_{nm} G_{mm}^\dag B_{nm}^\top
\ee 
The following are some examples of Generalized \Nystrom{} approximations.
\paragraph{Plain \Nystrom{} with various sampling schemes \cite{conf/icml/SmolaS00,conf/nips/WilliamsS00, Kumar:2012:SMN:2503308.2343678}}
For a realization  $s: \N \to \{1,\dots,n\}$ of a given sampling scheme, we choose $\Zm = S_m$ with $S_m^* = (K_{x_{s(1)}},\dots, K_{x_{s(m)}})^\top$ where $(x_i)_{i=1}^n$ is the training set. With such $\Zm$ we obtain $\tilde K_n =  K_{nm} (K_{mm})^\dag K_{nm}^\top$ and so Eq.~\eqref{eq:def-gen-nystrom} becomes exactly Eq.~\eqref{eq:repny}.
\paragraph{Reduced rank Plain \Nystrom{} \cite{Drineas:2005:NMA:1046920.1194916}}
Let $p \geq m$, $S_p $ as in the previous example, the linear operator associated to $p$ points of the dataset. Let $K_{pp} = S_p S_p^\top \in \R^{p\times p}$, that is $(K_{pp})_{ij} = K(x_i,x_j)$. Let $K_{pp} = \sum_{i=1}^p \sigma_i u_i u_i^\top$ its eigenvalue decomposition and $U_m = (u_1,\dots,u_m)$. Let $(K_{pp})_m = U_m^\top K_{pp} U_m$ be the $m$-rank approximation of $K_{pp}$.
We approximate this family by choosing $\Zm = U_m^\top S_p$, indeed we obtain $\tilde{\K} = K_{nm} U_m(U_m^\top K_{pp} U_m)^\dag U_m^\top K_{nm}^\top = K_{nm} (K_{pp})_m^\dag K_{nm}^\top$.

\paragraph{\Nystrom{} with sketching matrices \cite{gittens2013revisiting}}
We cover this family by choosing $\Zm = R_m S_n$, where $S_n$ is the same operator as in the plain \Nystrom{} case where we select all the points of the training set and $R_m$ a $m\times n$ sketching matrix. In this way we have
$\tilde{\K} =  {\K} R_m^* (R_m {\K} R_m^*)^\dag R_m {\K}$, that is exactly the SPSD sketching model.

\section{Probabilistic inequalities}
In this section we collect five main probabilistic inequalities needed in the proof of the main result.
We let $\rhox$ denote the marginal distribution of $\rho$ on $\X$ and $\rho(\cdot|x)$ the conditional distribution on $\R$ given $x\in \X$. Lemmas~\ref{lm:plain-nyst-error}, ~\ref{lemma:app-lev-scores-nystrom} and especially Proposition~\ref{lemma:emp-eff-dim} are new and of interest  in their own right.

The first result is essentially taken from \cite{caponnetto2007optimal}. 
\blm[Sample Error]\label{lm:sampling-error}
Under  Assumptions~\ref{as:exists-fh},~ \ref{as:noise} and~\ref{as:kerrho}, 
 for any $\delta > 0$, the following holds with probability $1- \delta$
\eqals{
\nor{(C+\la I)^{-1/2}(\Sn^* \yn  - \Cn f_\hh)} \leq 2 \left(\frac{M \sqrt{{\cal N}_\infty(\la)}}{n} + \sqrt{\frac{\sigma^2{\cal N}(\la)}{n}} \right)\log \frac{2}{\delta}
.}
\elm
\bpr
The proof is  given in \cite{caponnetto2007optimal} for bounded kernels and the slightly stronger condition $\int (e^\frac{|y - f_\hh(x)|}{M} - \frac{|y - f_\hh(x)|}{M} - 1) d\rho(y|x) \leq \sigma^2/M^2$ in place of Assumption \ref{as:noise}. More precisely, note that
$$(C+\la I)^{-1/2}(\Sn^* \yn  - \Cn f_\hh) = \frac{1}{n}\sum_{i=1}^n \zeta_i,$$
where $\zeta_1,\dots,\zeta_n$ are i.i.d. random variables, defined as $\zeta_i = (C+\la I)^{-1/2} K_{x_i} (y_i - f_\hh(x_i))$. For any $1\leq i \leq n$,
\eqals{ 
\mathbb{E} \zeta_i &= \int_{\X \times \R} (C+\la I)^{-1/2} K_{x_i} (y_i - f_\hh(x_i)) d\rho(x_i,y_i) \\
& = \int_X (C+\la I)^{-1/2} K_{x_i} \int_\R (y_i - f_\hh(x_i)) d\rho(y_i|x_i)d\rhox(x_i) = 0,
}
almost everywhere by Assumption~\ref{as:exists-fh} (see Step 3.2 of Thm. 4 in \cite{caponnetto2007optimal}). In the same way we have
\eqals{
\mathbb{E} \nor{\zeta_i}^p &= \int_{\X \times \R} \nor{(C+\la I)^{-1/2} K_{x_i} (y_i - f_\hh(x_i))}^p d\rho(x_i,y_i) \\
& = \int_X \nor{(C+\la I)^{-1/2} K_{x_i}}^p \int_\R |y_i - f_\hh(x_i)|^p d\rho(y_i|x_i)d\rhox(x_i) \\
& \leq \sup_{x \in \X} \nor{(C+\la I)^{-1/2} K_{x}}^{p-2} \int_X \nor{(C+\la I)^{-1/2} K_{x_i}}^2 \int_\R |y_i - f_\hh(x_i)|^p d\rho(y_i|x_i) d\rhox(x_i)\\
& \leq \frac{1}{2}p! \sqrt{\sigma^2\cal{N}(\la)}^2 (M\sqrt{{\cal N}_{\infty}(\la)})^{p-2}, 
}
where $\sup_{x \in \X} \nor{(C+\la I)^{-1/2} K_{x}} = \sqrt{{\cal N}_\infty(\la)}$ and $\int_X \nor{(C+\la I)^{-1/2} K_{x_i}}^2 = {\cal N}(\la)$ by Assumption~\ref{as:kerrho}, while the bound on the moments of $y - f(x)$ is given in Assumption~\ref{as:noise}.  Finally, to concentrate the sum of random vectors, we apply Prop.~\ref{prop:tail-vectors}.
%
\epr 
The next result is taken from  \cite{rudi2013sample}.
\blm\label{lm:bound-beta} Under Assumption~\ref{as:kerrho},
 for any $\delta \geq 0$ and  $\frac{9\kappa^2}{n}\log\frac{n}{\delta} \leq \la \leq \nor{C}$, the following inequality holds  with probability  at least $1-\delta$, 
\eqals{
\nor{(\Cn+\la I)^{-1/2} \C^{1/2}} \leq \nor{(\Cn+\la I)^{-1/2}(\C + \la I)^{1/2}} \leq 2.
}
\elm
\bpr
Lemma 7 of \cite{rudi2013sample} gives an
 the extended version of the above result. Our bound on $\la$ is scaled by $\kappa^2$ 
 because in \cite{rudi2013sample} it is assumed $\kappa \leq 1$.
\epr

\blm[plain \Nystrom{} approximation]\label{lm:plain-nyst-error}
Under Assumption~\ref{as:kerrho}, let $J$ be a partition of $\{1,\dots,n\}$ chosen uniformly at random from the partitions of cardinality $m$. Let $\la > 0$, for any $\delta > 0$, such that $m \geq 67 \log \frac{4\kappa^2}{\la \delta} \,\vee\, 5 {\cal N}_\infty(\la)\log\frac{4\kappa^2}{\la \delta}$,
the following holds with probability $1-\delta$
$$ \nor{(I - \Pm)(C+\la I)^{1/2}}^2 \leq 3\la,$$
where $\Pm$ is the projection operator on the subspace $\hh_m = \lspan{K_{x_j}~|~j \in J}$.
\elm
\bpr
Define the linear operator $\Cm: \hh \to \hh$, as $\Cm = \frac{1}{m} \sum_{j \in J} K_{x_j} \otimes K_{x_j}$. Now note that the range of $\Cm$ is exactly $\hh_m$. Therefore, by applying Prop.~\ref{lm:proj-bound} and \ref{prop:invAtoinvB}, we have that
\eqals{
\nor{(I - \Pm)\Cl^{1/2}}^2  \leq \la\nor{(\Cm +\la I)^{-1/2}\C^{1/2}}{}^2 \leq  \frac{\la}{1-\beta(\la)},
}
with $\beta(\la) = \la_{\max}\left(\Cl^{-1/2}(\C - \Cm)\Cl^{-1/2}\right)$.
To upperbound $\frac{\la}{1-\beta(\la)}$ we need an upperbound for $\beta(\la)$. Considering that, given the partition $J$, the random variables $\zeta_j = K_{x_j} \otimes K_{x_j}$ are i.i.d., then we can apply Prop.~\ref{prop:concentration-ClaC}, to obtain
$$ \beta(\la) \leq \frac{2w}{3 m} + \sqrt{\frac{2w {\cal N}_\infty(\la)}{m}}, $$
where $w = \log\frac{4 \tr(C)}{\la \delta}$ with probability $1-\delta$. Thus, by choosing $m \geq 67 w \vee 5 {\cal N}_\infty(\la) w$, we have that $\beta(\la) \leq 2/3$, that is
$$ \nor{(I - \Pm)\Cl^{1/2}}^2  \leq 3\la.$$ 
Finally, note that by definition $\tr(C) \leq \kappa^2$.
\epr

\blm[\Nystrom{} approximation for ALS selection method]\label{lemma:app-lev-scores-nystrom}
Let $(\hat l_i(t))_{i=1}^n$ be the collection of approximate leverage scores.
Let $\la > 0$ and $P_\la$ be defined as $P_\la(i) = \hat{l}_i(\la)/\sum_{j \in N}\hat{l}_j(\la)$ for any $i \in N$ with $N = \{1,\dots,n\}$. Let $\frak I = (i_1,\dots,i_m)$ be a collection of indices independently sampled with replacement from $N$ according to the probability distribution $P_\la$. Let $\Pm$ be the projection operator on the subspace $\hh_m = \lspan{K_{x_{j}}|j \in J}$ and $J$ be the subcollection of $\frak I$ with all the duplicates removed. Under Assumption~\ref{as:kerrho}, for any $\delta > 0$ the following holds with probability $1-2\delta$
$$
\nor{(I - \Pm)(C+\la I)^{1/2}} \leq 3\la,
$$
when the following conditions are satisfied:
\begin{enumerate}
\item there exists a $T \geq 1$ and a $\la_0 > 0$ such that $(\hat l_i(t))_{i=1}^n$ are $T$-approximate leverage scores for any $t \geq \la_0$ (see Def.~\ref{def:approx-lev-scores}),
\item $n \,\geq\, 1655\kappa^2 + 223\kappa^2\log\frac{2\kappa^2}{\delta}$,
\item $\la_0 \vee \frac{19\kappa^2}{n}\log\frac{2n}{\delta} \leq \la \leq \nor{C}{}$,
\item $m \,\geq\,  334 \log \frac{8n}{\delta}  \,\vee\, 78 T^2 {\cal N}(\la) \log \frac{8n}{\delta}$.
\end{enumerate}
\elm
\bpr Define $\tau = \delta/4$. Next, define the diagonal matrix $H \in \R^{n\times n}$ with $(H)_{ii} = 0$ when $P_\la(i) = 0$ and $(H)_{ii} = \frac{nq(i)}{mP_\la(i)}$ when $P_\la(i) > 0$, where $q(i)$ is the number of times the index $i$ is present in the collection $\frak I$. We have that
$$\Sn^* H \Sn = \frac{1}{m}\sum_{i=1}^n \frac{q(i)}{P_\la(i)} K_{x_i} \otimes K_{x_i} = \frac{1}{m}\sum_{j \in J} \frac{q(j)}{P_\la(j)} K_{x_j} \otimes K_{x_j}.$$
Now, considering that $\frac{q(j)}{P_\la(j)} > 0$ for any $j \in J$, thus $\ran{} \Sn^* H \Sn = \hh_m$. Therefore, by using Prop.~\ref{lm:proj-bound} and \ref{prop:invAtoinvB}, we exploit the fact that the range of $\Pm$ is the same of $\Sn^* H \Sn$, to obtain
\eqals{
\nor{(I - \Pm)(C+\la I)^{1/2}}^2  \leq \la\nor{(\Sn^* H \Sn +\la I)^{-1/2}(\C + \la I)^{1/2}}{}^2 \leq  \frac{\la}{1-\beta(\la)}, 
}
with $\beta(\la) = \la_{\max}\left(\Cl^{-1/2}(\C - \Sn^* H \Sn)\Cl^{-1/2}\right)$.
Considering that the function $(1-x)^{-1}$ is increasing on $-\infty < x < 1$, in order to bound $\la/(1-\beta(\la))$ we need an upperbound for $\beta(\la)$. Here we split $\beta(\la)$ in the following way,
$$
\beta(\la) \leq \underbrace{\la_{\max}\left(\Cl^{-1/2}(\C - \Cn)\Cl^{-1/2}\right)}_{\beta_1(\la)} + \underbrace{\la_{\max}\left(\Cl^{-1/2}(\Cn - \Sn^* H \Sn)\Cl^{-1/2}\right)}_{\beta_2(\la)}.
$$
Considering that $\Cn$ is the linear combination of independent random vectors, for the first term we can apply Prop.~\ref{prop:concentration-ClaC}, obtaining a bound of the form 
$$ \beta_1(\la) \leq \frac{2 w}{3 n} + \sqrt{\frac{2 w \kappa^2}{\lambda n}}, $$
with probability $1-\tau$, where $w = \log \frac{4\kappa^2}{\lambda\tau}$ (we used the fact that ${\cal N}_\infty(\la) \leq \kappa^2/\la$). Then, after dividing and multiplying by $\Cnl^{1/2}$, we split the second term $\beta_2(\la)$ as follows:
\begin{align*}
\beta_2(\la) &\leq \nor{\Cl^{-1/2}(\Cn - \Sn^* H \Sn)\Cl^{-1/2}}{} \\
& \leq \nor{\Cl^{-1/2}\Cnl^{1/2}\Cnl^{-1/2}(\Cn - \Sn^* H \Sn)\Cnl^{-1/2}\Cnl^{1/2}\Cl^{-1/2}}{} \\
& \leq \nor{\Cl^{-1/2}\Cnl^{1/2}}{}^2\nor{\Cnl^{-1/2}(\Cn - \Sn^* H \Sn)\Cnl^{-1/2}}{}.
\end{align*}
Let
\begin{align}
\beta_3(\la) = \nor{\Cnl^{-1/2}(\Cn - \Sn^* H \Sn)\Cnl^{-1/2}}{}  = \nor{\Cnl^{-1/2}\Sn^*(I - H)\Sn\Cnl^{-1/2}}{}.
\end{align}
Note that $\Sn\Cnl^{-1}\Sn^* = \K(\K + \la n I)^{-1}$ indeed $\Cnl^{-1} = (\Sn^*\Sn + \la I)^{-1}$ and $\K = n \Sn\Sn^*$. Therefore we have
$$\Sn\Cnl^{-1}\Sn^* = \Sn(\Sn^*\Sn + \la I)^{-1}\Sn^* = (\Sn\Sn^* + \la I)^{-1}\Sn\Sn^*  = (\K + \la n I)^{-1}\K.$$
Thus, if we let $U\Sigma U^\top$ be the eigendecomposition of $\K$, we have that $(\K + \la n I)^{-1}\K =  U(\Sigma+\la n I)^{-1}\Sigma U^\top$ and thus
 $\Sn\Cnl^{-1}\Sn^* = U(\Sigma+\la n I)^{-1}\Sigma U^\top$. In particular this implies that $\Sn\Cnl^{-1}\Sn^* = UQ_n^{1/2} Q_n^{1/2} U^\top$ with $Q_n = (\Sigma+\la n I)^{-1}\Sigma$. Therefore we have
$$
\beta_3(\la) = \nor{\Cnl^{-1/2}\Sn^*(I - H)\Sn\Cnl^{-1/2}}{} =  \nor{Q_n^{1/2}U^\top(I - H)UQ_n^{1/2}}{},$$
where we used twice the fact that $\nor{A B A^*} = \nor{(A^*A)^{1/2} B (A^*A)^{1/2}}$ for any bounded linear operators $A, B$.

Consider the matrix $A = Q_n^{1/2}U^\top$ and let $a_i$ be the $i$-th column of $A$, and $e_i$ be the $i$-th canonical basis vector for each $i \in N$. We prove that $\nor{a_i}{}^2 = l_i(\la)$, the true leverage score, since
$$\nor{a_i}{}^2 = \nor{Q_n^{1/2}U^\top e_i}{}^2 = e_i^\top U Q_n U^\top e_i = ((\K + \la n I)^{-1}\K)_{ii} = l_i(\la).$$

Noting that $\sum_{k = 1}^n \frac{q(k)}{P_\la(k)} a_k a_k^\top = \sum_{i = \frak I} \frac{1}{P_\la(i)} a_i a_i^\top$, we have
$$\beta_3(\la) = \nor{AA^\top - \frac{1}{m}\sum_{i \in \frak I} \frac{1}{P_\la(i)} a_i a_i^\top}{}.$$ 
Moreover, by the $T$-approximation property of the approximate leverage scores (see Def.~\ref{def:approx-lev-scores}), we have that for all $i \in \{1,\dots,n\}$, when $\la \geq \la_0$, the following holds with probability $1-\delta$
$$P_\la(i) = \frac{\hat l_i(\la)}{\sum_j \hat l_j(\la)} \geq T^{-2} \frac{l_i(\la)}{\sum_j l_j(\la)} = T^{-2} \frac{\nor{a_i}{}^2}{\tr AA^\top}.$$
Then, we can apply Prop.~\ref{prop:conc-subsample}, so that, after a union bound, we 
 obtain the following inequality  with probability $1-\delta-\tau$:
$$\beta_3(\la) \leq \frac{2 \nor{A}{}^2 \log \frac{2n}{\tau}}{3m} + 
\sqrt{\frac{2\nor{A}{}^2 T^2\tr AA^\top \log\frac{2n}{\tau}}{m}} \leq \frac{2 \log \frac{2n}{\tau}}{3m} + 
\sqrt{\frac{2 T^2 \hat{\cal N}(\la) \log\frac{2n}{\tau}}{m}},$$
where the last step follows from 
$\nor{A}{}^2 = \nor{(\K + \la n I)^{-1}\K}{} \leq 1$ and 
$\tr(AA^\top) = \tr(\Cnl^{-1}\Cn) := \hat{\cal N}(\la)$.
Applying Proposition~\ref{lemma:emp-eff-dim}, we have that $\hat{\cal N}(\la) \leq 1.3{\cal N}(\la)$ with probability $1-\tau$, when $\frac{19\kappa^2}{n}\log\frac{n}{4\tau} \leq \la \leq \nor{C}{}$ and $n \geq 405\kappa^2 \vee 67\kappa^2\log\frac{\kappa^2}{2\tau}$. Thus, by taking a union bound again, we have
$$ \beta_3(\la) \leq \frac{2 \log \frac{2n}{\tau}}{3m} + 
\sqrt{\frac{5.3 T^2 {\cal N}(\la) \log\frac{2n}{\tau}}{m}},$$
with probability $1-2\tau-\delta$ when $\la_0 \vee \frac{19\kappa^2}{n}\log\frac{n}{\delta} \leq \la \leq \nor{C}{}$ and $n \geq 405\kappa^2 \vee 67\kappa^2\log\frac{2\kappa^2}{\delta}$.
The last step is to bound $\nor{\Cl^{-1/2}\Cnl^{1/2}}{}^2$, as follows
$$\nor{\Cl^{-1/2}\Cnl^{1/2}}{}^2 = \nor{\Cl^{-1/2}\Cnl\Cl^{-1/2}}{} = \nor{I + \Cl^{-1/2}(\Cn - \C)\Cl^{-1/2}}{} \leq 1 + \eta,$$
with $\eta = \nor{\Cl^{-1/2}(\Cn - \C)\Cl^{-1/2}}{}$. Note that, by applying Prop.~\ref{prop:concentration-ClaC} we have that
$\eta \leq \frac{2(\kappa^2+\la)\theta}{3\la n} + \sqrt{\frac{2\kappa^2\theta}{3\la n}}$ with probability $1-\tau$ and $\theta = \log\frac{8 \kappa^2}{\la \tau}$.
Finally, by collecting the above results and taking a union bound we have
$$ \beta(\la) \leq  \frac{2 w}{3 n} + \sqrt{\frac{2 w \kappa^2}{\la n}} + (1+\eta)\left(\frac{2 \log \frac{2n}{\tau}}{3m} + 
\sqrt{\frac{5.3 T^2 {\cal N}(\la) \log\frac{2n}{\tau}}{m}}\right),$$
with probability $1-4\tau-\delta = 1-2\delta$ when $\la_0 \vee \frac{19\kappa^2}{n}\log\frac{n}{\delta} \leq \la \leq \nor{C}{}$ and $n \geq 405\kappa^2 \vee 67\kappa^2\log\frac{2\kappa^2}{\delta}$. Note that, if we select $n \geq 405\kappa^2 \vee 223\kappa^2\log\frac{2\kappa^2}{\delta}$, $m \geq 334 \log \frac{8n}{\delta}$, $\la_0 \vee \frac{19\kappa^2}{n}\log\frac{2n}{\delta} \leq \la \leq \nor{C}{}$ and $\frac{78 T^2 {\cal N}(\la) \log \frac{8n}{\delta}}{m} \leq 1$ the conditions are satisfied and we have $\beta(\la) \leq 2/3$, so that
$$
\nor{(I - \Pm)(\C+\la I)^{1/2}}{}^2 \leq 3\la,
$$
with probability $1-2\delta$.
\epr

\bp[Empirical Effective Dimension]\label{lemma:emp-eff-dim} Let $\hat{\cal N}(\la) = \tr \Cn\Cnl^{-1}$. Under the Assumption~\ref{as:kerrho}, for any $\delta > 0$ and $n \geq 405\kappa^2 \vee 67\kappa^2\log\frac{6\kappa^2}{\delta}$, if $\frac{19\kappa^2}{n}\log\frac{n}{4\delta} \leq \la \leq \nor{C}{}$, then the following holds with probability $1-\delta$,
$$
\frac{|\hat{\cal N}(\la) - {\cal N}(\la)|}{{\cal N}(\la)} \leq 4.5 q + (1 + 9 q) \sqrt{\frac{3q}{\cal{N}(\la)}} + \frac{q + 13.5 q^2}{\cal{N}(\la)} \leq 1.65,
$$
with $q = \frac{4\kappa^2 \log \frac{6}{\delta}}{3 \la n}$.
\ep
\bpr
Let $\tau = \delta/3$. Define $B_n = \Cl^{-1/2}(\C - \Cn)\Cl^{-1/2}$. Choosing $\la$ in the range $\frac{19\kappa^2}{n}\log\frac{n}{4\tau} \leq \la \leq \nor{C}{}$, Prop.~\ref{prop:concentration-ClaC} assures that $\la_{\max}(B_n) \leq 1/3$ with probability $1-\tau$.
Then, using the fact that $\Cnl^{-1} = \Cl^{-1/2}(I-B_n)^{-1}\Cl^{-1/2}$ (see the proof of Prop.~\ref{prop:invAtoinvB}) we have
\begin{align*}
|\hat{\cal N}(\la) - {\cal N}(\la)|  & =  |\tr \Cnl^{-1}\Cn  - \C\Cl^{-1} = \la \tr \Cnl^{-1}(\Cn - \C)\Cl^{-1}| \\
{} & = |\la \tr \Cl^{-1/2}\left(I - B_n\right)^{-1}\Cl^{-1/2} (\Cn - \C)\Cl^{-1/2}\Cl^{-1/2}|\\
{} & =  |\la \tr \Cl^{-1/2}\left(I - B_n\right)^{-1}B_n\Cl^{-1/2}|.
\end{align*}
Considering that for any symmetric linear operator $X: \hh \to \hh$ the following identity holds $$(I - X)^{-1}X = X + X(I-X)^{-1}X,$$ when $\la_{\max}(X) \leq 1$, we have 
\begin{align*}
\la |\tr \Cl^{-1/2}\left(I - B_n\right)^{-1}B_n\Cl^{-1/2}|  & \leq  \underbrace{\la |\tr \Cl^{-1/2}B_n\Cl^{-1/2}|}_{A} \\
& \quad\quad + \underbrace{\la |\tr \Cl^{-1/2}B_n\left(I - B_n\right)^{-1}B_n\Cl^{-1/2}|}_{B}.
\end{align*} 

To find an upperbound for $A$ define the i.i.d. random variables $\eta_i = \scal{K_{x_i}}{\la\Cl^{-2} K_{x_i}} \in \R$ with $i \in \{1,\dots,n\}$. By linearity of the trace and the expectation, we have $M = \mathbb{E} \eta_1  = \mathbb{E} \scal{K_{x_i}}{\la\Cl^{-2} K_{x_i}} = \mathbb{E} \tr (\la \Cl^{-2} K_{x_1}\otimes K_{x_1}) = \la \tr (\Cl^{-2}\C)$. Therefore,
\begin{align*}
\la |\tr \Cl^{-1/2}B_n\Cl^{-1/2}| = \left|M - \frac{1}{n}\sum_{i=1}^n \eta_i\right|,
\end{align*}
and we can apply the Bernstein inequality (Prop.~\ref{prop:tail-bern}) with 
$$|M - \eta_1| \leq \la\nor{\Cl^{-2}}{}\nor{K_{x_1}}{}^2 + M \leq \frac{\kappa^2}{\la} + M \leq  \frac{2\kappa^2}{\la} = L,$$
$$\mathbb{E}(\eta_1 - M)^2 = \mathbb{E}\eta_1^2 - M^2 \leq \mathbb{E}\eta_1^2 \leq L M = \sigma^2.$$
An upperbound for $M$ is $M = \tr (\la \Cl^{-2}C) = \tr((I - \Cl^{-1}C)\Cl^{-1}C) \leq \cal{N}(\la)$. Thus, we have
$$ \la |\tr \Cl^{-1/2}B_n\Cl^{-1/2}| \leq \frac{4\kappa^2 \log \frac{2}{\tau}}{3\la n} + \sqrt{\frac{4\kappa^2 \mathcal{N}(\la) \log \frac{2}{\tau}}{\la n}},$$
with probability $1-\tau$.

To find an upperbound for $B$, let $\cal L$ be the space of Hilbert-Schmidt operators on $\hh$. $\cal L$ is a Hilbert space with scalar product $\scal{U}{V}_{HS} = \tr{(UV^*)}$ for all $U, V \in \cal L$. Next, note that $B = \nor{Q}_{HS}^2$ where $Q = \la^{1/2}\Cl^{-1/2}B_n\left(I - B_n\right)^{-1/2}$, moreover
 $$\nor{Q}_{HS}^2 \leq \nor{\la^{1/2}\Cl^{-1/2}}{}^2\nor{B_n}_{HS}^2\nor{\left(I - B_n\right)^{-1/2}}{}^2 \leq 1.5 \nor{B_n}_{HS}^2,$$
since $\nor{(I - B_n)^{-1/2}}{}^2 = (1 - \la_{\max}(B_n))^{-1} \leq 3/2$ and $(1-\sigma)^{-1}$ is increasing and positive on $[-\infty,1)$.

To find a bound for $\nor{B_n}_{HS}$ consider that $B_n = T - \frac{1}{n}\sum_{i=1}^n \zeta_i$ where $\zeta_i$ are i.i.d. random operators defined as $\zeta_i = \Cl^{-1/2} (K_{x_i} \otimes  K_{x_i}) \Cl^{-1/2} \in \cal L$ for all $i \in \{1,\dots, n\}$, and $T = \mathbb{E} \zeta_1 = \Cl^{-1}C \in \cal L$.
Then we can apply the Bernstein's inequality for random vectors on a Hilbert space (Prop.~\ref{prop:tail-vectors}), with the following $L$ and $\sigma^2$: 
$$\nor{T - \zeta_1}_{HS} \leq \nor{\Cl^{-1/2}}{}^2 \nor{K_{x_1}}_{\hh}^2 + \nor{T}_{HS} \leq \frac{ \kappa^2}{\la} + \nor{T}_{HS} \leq \frac{2\kappa^2}{\la} = L,$$
$$\mathbb{E} \nor{\zeta_1 - T}{}^2 = \mathbb{E} \tr(\zeta_1^2 - T^2) \leq \mathbb{E} \tr(\zeta_1^2) \leq L \mathbb{E} \tr(\zeta_1) = \sigma^2,$$
where $\nor{T}_{HS} \leq \mathbb{E}\tr(\zeta_1)  = \mathcal{N}(\la)$, obtaining
$$\nor{B_n}_{HS} \leq \frac{4\kappa^2 \log \frac{2}{\tau}}{\la n} +  \sqrt{\frac{4\kappa^2 \mathcal{N}(\la) \log \frac{2}{\tau}}{\la n}},$$
with probability $1-\tau$.
Then, by taking a union bound for the three events we have
$$
|\hat{\cal N}(\la) - {\cal N}(\la)| \leq q + \sqrt{3 q \mathcal{N}(\la)} + 1.5\left(3q + \sqrt{3 q \mathcal{N}(\la)}\right)^2,
$$
with $q = \frac{4\kappa^2 \log \frac{6}{\delta}}{3\la n}$, and with probability $1-\delta$.
Finally, if the second assumption on $\la$ holds, then we have $q \leq 4/57$. Noting that $n \geq 405\kappa^2$, and that ${\cal N}(\la) \geq \nor{C\Cl^{-1}}{} = \frac{\nor{C}{}}{\nor{C}{}+\la} \geq 1/2$, we have that
$$|\hat{\cal N}(\la) - {\cal N}(\la)| \leq \left(\frac{q}{3{\cal N}(\la)} + \sqrt{\frac{ q}{\mathcal{N}(\la)}} + 1.5\left(\frac{q}{\sqrt{{\cal N}(\la)}} + \sqrt{ q}\right)^2\right){\cal N}(\la) \leq 1.65 {\cal N}(\la).$$
\epr

\section{Proofs of main theorem}
A key step to derive the proof of Theorem~\ref{thm:opt-rates-NyKRLS} is the error decomposition given by the following theorem, together with the probabilistic inequalities in the previous section. 

\bt[Error decomposition for KRLS+Ny]\label{thm:nys-err-dec}
Under Assumptions~\ref{as:exists-fh},~\ref{as:kerrho},~\ref{as:source}, let $v = \min(s,1/2)$ and $\hat{f}_{\la,m}$ a KRLS + generalized \Nystrom{} solution as in Eq.~\eqref{eq:def-gen-nystrom}. Then for any $\la, m > 0$ the error is bounded by
\eqal{\label{eq:nys-err-dec-thesis}
\left|\EE(\hat{f}_{\la,m}) - \EE(f_\hh)\right|^{1/2} \leq q (\underbrace{{\cal S}(\la, n)}_{\textrm{Sample error}} \; + \,{\underbrace{{\cal C}(m)^{1/2 + v}}_{\textrm{Computational error}}}\, + \underbrace{\la^{1/2+v}}_{\textrm{Approximation error}}),
}
where ${\cal S}(\la, n) = \nor{(C+\la I)^{-1/2}(\Sn^* \yn - \Cn f_\hh)}$ and ${\cal C}(m) = \nor{(I - P_m)(C+\la I)^{1/2}}^2$ with $P_m = \Zm^*(\Zm\Zm^*)^\dag\Zm$. Moreover $q = R( \beta^2 \vee (1 + \theta\beta))$, $\beta = \nor{(\Cn + \la I)^{-1/2} (\C + \la I)^{1/2}}$, $\theta = \nor{(\Cn + \la I)^{1/2}(C + \la I)^{-1/2}}$.
\et
\bpr
Let $\Cl = \C + \la I$ and $\Cnl = \Cn + \la I$ for any $\la > 0$. Let $\hat{f}_{\la,m}$ as in Eq.~\eqref{eq:def-gen-nystrom}. By Lemma~\ref{lm:char0-flam}, Lemma~\ref{lm:char1-flam} and Lemma~\ref{lm:flm-rewriting} we know that $\hat{f}_{\la,m}$ is characterized by $\hat f_{\la,m} = g_{\la m}(\Cn)S_n^*\yn$ with $g_{\la, m}(\Cn) = V(V^*\Cn V + \la I)^{-1} V^*$. By using the fact that
$\EE(f) - \EE(f_\hh) =  \nor{C^{1/2}(f - f_\hh)}^2_\hh$ for any $f \in \hh$ (see Prop. 1 Point 3 of \cite{caponnetto2007optimal}), we have
\eqals{
|\EE(\hat f_{\la, m}) - \EE(f_\hh)|^{1/2} &= \nor{C^{1/2}(\hat f_{\la, m} - f_\hh)}_\hh
 = \nor{\C^{1/2}(g_{\la, m}(\Cn) \Sn^* \yn - f_\hh)}_\hh \\
& =  \nor{\C^{1/2}( g_{\la, m}(\Cn) \Sn^*(\yn - \Sn f_\hh + \Sn f_\hh) - f_\hh)}_\hh \\
& \leq  \underbrace{\nor{\C^{1/2}g_{\la, m}(\Cn) \Sn^*(\yn - \Sn f_\hh)}_\hh}_{A} +  \underbrace{\nor{\C^{1/2}(I - g_{\la, m}(\Cn) \Cn) f_\hh}_\hh}_{B}.
}
\textbf{Bound for the term A} 
Multiplying and dividing by  $\Cnl^{1/2}$ and $\Cl^{1/2}$ we have
\eqals{
A &\leq \nor{\C^{1/2}\Cnl^{-1/2}}\nor{\Cnl^{1/2}g_{\la, m}(\Cn) \Cnl^{1/2}}\nor{\Cnl^{-1/2}\Cl^{1/2}}\nor{\Cl^{-1/2}\Sn^*({\widehat y_n} - \Sn f_\hh)}_\hh 
 \leq \beta^2 \, {\cal S}(\la, n),
}
where the last step is due to Lemma~\ref{lm:simpl-glm} and the fact that
$$ \nor{\C^{1/2}\Cnl^{-1/2}} \leq \nor{\C^{1/2}\Cl^{-1/2}}\nor{\Cl^{1/2}\Cnl^{-1/2}} \leq \nor{\Cl^{1/2}\Cnl^{-1/2}}.$$

\textbf{Bound for the term B} 
Noting that $g_{\la, m}(\Cn) \Cnl VV^* = VV^*$, we have
\eqals{
I - g_{\la, m}(\Cn) \Cn &= I - g_{\la, m}(\Cn) \Cnl + \la g_{\la, m}(\Cn) \\
& = I - g_{\la, m}(\Cn) \Cnl VV^* - g_{\la, m}(\Cn) \Cnl(I - VV^*)  + \la g_{\la, m}(\Cn)\\
& = (I - VV^*) + \la g_{\la, m}(\Cn) - g_{\la, m}(\Cn)\Cnl(I - VV^*). 
}
Therefore, noting that by Ass.~\ref{as:source} we have $\nor{\Cl^{-v} f_\hh}_\hh \leq \nor{\Cl^{-s} f_\hh}_\hh \leq \nor{\C^{-s} f_\hh}_\hh \leq R$, then, by reasoning as in A, we have     
\eqals{
B & \leq \nor{\C^{1/2}(I - g_{\la, m}(\Cn) \Cn) \Cl^v}\nor{\Cl^{-v} f_\hh}_\hh \\
& \leq R\nor{\C^{1/2}\Cl^{-1/2}}\nor{\Cl^{1/2} (I-VV^*) \Cl^v} +  R\la \nor{\C^{1/2} \Cnl^{-1/2}}\nor{\Cnl^{1/2} g_{\la, m}(\Cn) \Cl^v}\\
& \quad + R\nor{\C^{1/2} \Cnl^{-1/2}}\nor{\Cnl^{1/2} g_{\la, m}(\Cn)\Cnl^{1/2}}\nor{\Cnl^{1/2}\Cl^{-1/2}}\nor{\Cl^{1/2}(I - VV^*)\Cl^v} \\
& \leq  R(1 + \beta\theta)\underbrace{\nor{\Cl^{1/2}(I - VV^*)\Cl^v}}_{B.1} + R \beta \underbrace{\la\nor{\Cnl^{1/2} g_{\la, m}(\Cn) \Cl^v}}_{B.2},
}
where in the second step we applied the decomposition of $I - g_{\la m}(\Cn) \Cn$. 

\textbf{Bound for the term B.1} Since $VV^*$ is a projection operator, we have that $(I - VV^*) = (I - VV^*)^s$, for any $s > 0$, therefore
$$B.1 = \nor{\Cl^{1/2} (I - VV^*)^2\Cl^v}  \leq \nor{\Cl^{1/2} (I - VV^*)}\nor{(I - VV^*)\Cl^v}.$$
By applying Cordes inequality (Prop.~\ref{prop:cordes}) to $\nor{(I - VV^*)\Cl^v}$ we have,
\eqals{\nor{(I - VV^*)\Cl^v} &= \nor{(I - VV^*)^{2v}\Cl^{\frac{1}{2}2v}} = \nor{(I - VV^*)\Cl^{1/2}}^{2v}.
}
\textbf{Bound for the term B.2} We have
\eqals{
B.2 & \leq \la\nor{\Cnl^{1/2}g_{\la, m}(\Cn)\Cnl^{v}}\nor{\Cnl^{-v}\Cl^v} \\
& \leq \la\nor{\Cnl^{1/2}g_{\la, m}(\Cn)\Cnl^{v}}\nor{\Cnl^{-1/2}\Cl^{1/2}}^{2v}\\
& \leq \beta^{2v} \la \nor{(V^*\Cnl V)^{1/2}(V^*\Cnl V)^{-1}(V^*\Cnl V)^{v}}\\
& = \beta^{2v}\la \nor{(V^*\Cn V + \la I)^{-(1/2-v)}} \leq \beta\la^{1/2+v},
}
where the first step is obtained multipling and dividing by $\Cnl^v$, the second step by applying Cordes inequality (see Prop.~\ref{prop:cordes}), the third step by Prop.~\ref{prop:hansen-ineq}. 
\epr 

\bp[Bounds for plain and ALS \Nystrom{}]\label{prop:bounds-plain-appr}
For any $\delta > 0$, let $n \,\geq\, 1655\kappa^2 + 223\kappa^2\log\frac{6\kappa^2}{\delta}$, let $\frac{19\kappa^2}{n}\log\frac{6n}{\delta} \leq \la \leq \nor{C}{}$ and define
\eqals{
{\cal C}_{\rm pl}(m) &= \min \left\{t > 0 ~\middle|~ (67 \vee 5 {\cal N}_\infty(t))\log\frac{12\kappa^2}{t \delta} \leq m \right\}, \\
{\cal C}_{\rm ALS}(m) & = \min \left\{\frac{19\kappa^2}{n}\log\frac{12n}{\delta} \leq t \leq \nor{C}{} ~\middle|~ 78 T^2 {\cal N}(t) \log \frac{48n}{\delta} \leq m \right\}.
}
Under the assumptions of Thm.~\ref{thm:nys-err-dec} and Assumption~\ref{as:noise},~\ref{as:kerrho}, if one of the following two conditions hold
\begin{enumerate}
\item plain \Nystrom{} is used, 
\item ALS \Nystrom{} is used with
\begin{enumerate}
\item $T$-approximate leverage scores, for any $t \geq \frac{19\kappa^2}{n}\log\frac{12n}{\delta}$ (see Def.~\ref{def:approx-lev-scores}),
\item resampling probabilities $P_t$ where $t = {\cal C}_\textrm{ALS}(m)$ (see Sect.~\ref{sect:krls-nyst}),
\item $m \geq   334 \log \frac{48n}{\delta}$,
\end{enumerate}
\end{enumerate}
then the following holds with probability $1-\delta$
\eqal{
\left|{\cal E}(\hat f_{\la, m}) - {\cal E}(f_{\hh})\right|^{1/2} \leq 6R\left(\frac{M \sqrt{{\cal N}_\infty(\la)}}{n} + \sqrt{\frac{\sigma^2 {\cal N}(\la)}{n}}\right)\log \frac{6}{\delta} +  3R {\cal C}(m)^{1/2 + v} + 3R\la^{1/2+v}
}
where ${\cal C}(m) = {\cal C}_{\rm pl}(m)$ in case of plain \Nystrom{} and ${\cal C}(m) = {\cal C}_{\rm ALS}(m)$ in case of ALS \Nystrom{}.
\ep
\bpr In order to get explicit bounds from Thm.~\ref{thm:nys-err-dec}, we have to control four quantities that are
$\beta, \theta, {\cal S}(\la,n)$ and ${\cal C}(m)$. In the following we bound such quantities in probability and then take a union bound. Let $\tau = \delta/3$.
We can control both $\beta$ and $\theta$, by bounding $b(\la) = \nor{\Cl^{-1/2}(\Cn - C)\Cl^{-1/2}}$. Indeed, by Prop.~\ref{prop:invAtoinvB}, we have that
$\beta \leq 1/(1-b(\la))$, while 
$$\theta^2 = \nor{\Cl^{-1/2}\Cnl \Cl^{-1/2}} = \nor{I + \Cl^{-1/2}(\Cn - C)\Cl^{-1/2}} \leq 1 + b(\la).$$
Exploiting Prop.~\ref{prop:concentration-ClaC}, with the fact that ${\cal N}(\la) \leq {\cal N}_\infty(\la) \leq \frac{\kappa^2}{\la}$ and $\tr C \leq \kappa^2$, we have that
$b(\la) \leq \frac{2(\kappa^2+\la)w}{3\la n} + \sqrt{\frac{2w\kappa^2}{\la n}}$ for $w = \log\frac{4\kappa^2}{\tau \la}$ with probability $1-\tau$. Simple computations show that with $n$ and $\la$ as in the statement of this corollary, we have $b(\la) \leq 1/3$. Therefore $\beta \leq 1.5$, while $\theta \leq 1.16$ and $q = R(\beta^2 \vee (1 + \theta\beta)) < 2.75 R$ with probability $1-\tau$. Next, we bound ${\cal S}(\la, n)$. Here we exploit Lemma~\ref{lm:sampling-error} which gives, with probability $1-\tau$,
\eqals{
{\cal S}(\la, n) \leq 2 \left(\frac{M \sqrt{{\cal N}_\infty(\la)}}{n} + \sqrt{\frac{\sigma^2{\cal N}(\la)}{n}} \right)\log \frac{2}{\tau}.
}
To bound ${\cal C}(m)$ for plain \Nystrom{}, Lemma~\ref{lm:plain-nyst-error} gives ${\cal C}(m) \leq 3t$ with probability $1-\tau$, for a $t > 0$ such that $(67 \vee 5 {\cal N}_\infty(t))\log\frac{4\kappa^2}{t \tau} \leq m$. In particular, we choose $t = {\cal C}_{\rm pl}(m)$ to satisfy the condition.
Next we bound ${\cal C}(m)$ for ALS \Nystrom{}. Using Lemma~\ref{lemma:app-lev-scores-nystrom} with $\la_0 = \frac{19\kappa^2}{n}\log\frac{2n}{\tau}$, we have ${\cal C}(m) \leq 3t$ with probability $1-\tau$ under some conditions on $t, m, n$, on the approximate leverage scores and on the resampling probability. Here again the requirement on $n$ is satisfied by the hypotesis on $n$ of this proposition, while the condition on the approximate leverage scores and on the resampling probabilities are satisfied by conditions (a), (b) of this proposition. The remaining two conditions are $\frac{19\kappa^2}{n}\log\frac{4n}{\tau} \leq t \leq \nor{C}{}$ and $(334 \vee 78 T^2 {\cal N}(t)) \log \frac{16n}{\tau} \leq m$. They are satisfied by choosing $t = {\cal C}_{\rm ALS}(m)$ and by assuming that $m \geq 334 \log \frac{16n}{\tau}$. Finally, the proposition is obtained by substituting each of the four quantities $\beta, \theta, {\cal S}(\la, n), {\cal C}(m)$ with the corresponding upperbounds in Eq.~\eqref{eq:nys-err-dec-thesis}, and by taking the union bounds on the associated events.
\epr

\bpr[Proof of Theorem~\ref{thm:opt-rates-NyKRLS}]
By exploiting the results of Prop.~\ref{prop:bounds-plain-appr}, obtained from the error decomposition of Thm.~\ref{thm:nys-err-dec} we have that
\eqal{\label{eq:eq-pr-thm1}
\left|{\cal E}(\hat f_{\la, m}) - {\cal E}(f_{\hh})\right|^{1/2} \leq 6R\left(\frac{M \sqrt{{\cal N}_\infty(\la)}}{n} + \sqrt{\frac{\sigma^2 {\cal N}(\la)}{n}}\right)\log \frac{6}{\delta} +  3R {\cal C}(m)^{1/2 + v} + 3R\la^{1/2+v}
}
with probability $1-\delta$, under conditions on $\la, m, n$, on the resampling probabilities and on the approximate leverage scores.
The last is satisfied by condition (a) in this theorem. The conditions on $\la, n$ are $n \geq 1655\kappa^2 + 223\kappa^2\log\frac{6\kappa^2}{\delta}$ and $\frac{19\kappa^2}{n}\log\frac{12n}{\delta} \leq \la \leq \nor{C}{}$. 
If we assume that $n \geq  1655\kappa^2 + 223\kappa^2\log\frac{6\kappa^2}{\delta} + \left(\frac{38p}{\nor{C}} \log \frac{114\kappa^2 p}{\nor{C}\delta} \right)^p$ we satisfy the condition on $n$ and at the same time we are sure that $\la = \nor{C} n^{-1/(2v+\gamma + 1)}$ satisfies the condition on $\la$.
In the plain \Nystrom{} case, if we assume that $m \geq 67 \log \frac{12\kappa^2}{\la \delta} + 5 {\cal N}_\infty(\la)\log\frac{12\kappa^2}{\la \delta}$, then ${\cal C}(m) = {\cal C}_{\rm pl}(m) \leq \la$. In the ALS \Nystrom{} case, if we assume that $m \geq (334 \vee 78 T^2 {\cal N}(\la)) \log \frac{48n}{\delta}$ the condition on $m$ is satisfied, then ${\cal C}(m) = {\cal C}_{\rm ALS}(m) \leq \la$, moreover the conditions on the resampling probabilities is satisfied by condition (b) of this theorem. Therefore, by setting $\la = \nor{C}n^{-1/(2v+\gamma + 1)}$ in Eq.~\eqref{eq:eq-pr-thm1} and considering that ${\cal N}_\infty(\la) \leq \kappa^2\la^{-1}$ we easily obtain the result of this theorem.
\epr

The following lemma is a technical result needed in the error decomposition (Thm.~\ref{thm:nys-err-dec}).
\blm\label{lm:simpl-glm}
For any $\la > 0$, let $V$ be such that $V^* V = I$ and $\Cn$ be a positive self-adjoint operator. Then,  the following holds, 
$$\nor{(\Cn + \la I)^{1/2}V(V^*\Cn V + \la I)^{-1} V^* (\Cn + \la I)^{1/2}} \leq 1.$$
\elm
\bpr 
Let $\Cnl = \Cn + \la I$ and $g_{\la m}(\Cn) = V(V^*\Cn V + \la I)^{-1} V^*$, then 
\eqals{
\nor{\Cnl^{1/2}g_{\la m}(\Cn) \Cnl^{1/2}}^2 &= \nor{\Cnl^{1/2}g_{\la m}(\Cn)\Cnl g_{\la m}(\Cn) \Cnl^{1/2}}\\
 &= \nor{\Cnl^{1/2}V(V^*\Cnl V)^{-1} (V^*\Cnl V)(V^*\Cnl V)^{-1} V^*\Cnl^{1/2}} \\
 &= \nor{\Cnl^{1/2}g_{\la m}(\Cn)\Cnl^{1/2}},
}
and therefore the only possible values for $\nor{\Cnl^{1/2}g_{\la m}(\Cn)\Cnl^{1/2}}$ are $0$ or $1$.
\epr

\section{Auxiliary results}

\bp\label{lm:proj-bound}
Let $\hh, \kk, {\cal F}$ three separable Hilbert spaces, let $Z: \hh \to \kk$ be a bounded linear operator and let $P$ be a projection operator on $\hh$ such that $\ran P = \overline{\ran{Z^*}}$. Then for any bounded linear operator $F: {\cal F} \to \hh$ and any $\la > 0$ we have
$$ \nor{(I - P)X} \leq \la^{1/2} \nor{(Z^*Z + \la I)^{-1/2} X} .$$
\ep
\bpr First of all note that $\la(Z^*Z + \la I)^{-1} = I - Z^*(ZZ^* +\la I)^{-1}Z$, that $Z = Z P$ and that $\nor{Z^*(ZZ^* +\la I)^{-1}Z} \leq 1$ for any $\la > 0$. Then for any $v \in \hh $ we have
\eqals{
\scal{v}{Z^*(ZZ^* +\la I)^{-1}Z v} &= \scal{v}{PZ^*(ZZ^* +\la I)^{-1}ZP v} = \nor{(ZZ^* +\la I)^{-1/2}ZP v}^2 \\
& \leq \nor{(ZZ^* +\la I)^{-1/2}Z}^2\nor{P v}^2 \leq \nor{P v}^2 = \scal{v}{ P v}
}
therefore $P - Z^*(ZZ^* +\la I)^{-1}Z$ is a positive operator, and $(I - Z^*(ZZ^* +\la I)^{-1}Z) - (I-P)$ too. Now we can apply Prop.~\ref{prop:lop-prop}.
\epr

\bp[Cordes Inequality \cite{fujii1993norm}]\label{prop:cordes}
Let $A, B$ two positive semidefinite bounded linear operators on a separable Hilbert space. Then
$$\nor{A^s B^s} \leq \nor{A B}^s \quad \textrm{when } 0 \leq s \leq 1 .$$ 
\ep

\bp\label{prop:lop-prop}
Let $\hh, \kk, {\cal F}, {\cal G}$ be three separable Hilbert spaces and let $X: \hh \to \kk$ and $Y: \hh \to {\cal F}$ be two bounded linear operators. For any bounded linear operator $Z: {\cal G} \to \hh$, if $Y^*Y - X^*X$ is a positive self-adjoint operator then $\nor{X Z}{} \leq \nor{Y Z}{}$. 
\ep
\bpr
If $Y^*Y - X^*X$ is a positive operator then $Z^*(Y^*Y - X^*X)Z$ is positive too. Thus for all $f \in \hh$ we have that $\scal{f}{(Q-P)f}{} \geq 0$, where $Q = Z^*Y^*YZ$ and $P = Z^*X^*XZ$. Thus, by linearity of the inner product, we have
$$ \nor{Q}{} = \sup_{f\in {\cal G}}\scal{f}{Q f} = \sup_{f\in {\cal G}} \left\{\scal{f}{P f} + \scal{f}{(Q-P)f}\right\} \geq \sup_{f\in {\cal G}} \scal{f}{Pf} = \nor{P}{}.$$\\
\epr

\bp\label{prop:hansen-ineq}
Let $\hh, \kk$ be two separable Hilbert spaces, let $A: \hh \to \hh$ be a positive linear operator, $V: \hh \to \kk$ a partial isometry and $B: \kk \to \kk$ a bounded operator. Then
$\nor{A^r V B V^* A^s} \leq \nor{(V^* A V)^{r}B(V^* A V)^{s}}$, for all $0 \leq r,s \leq 1/2$.
\ep
\bpr By Hansen's inequality (see \cite{hansen1980operator}) we know that $(V^* A V)^{2t} - V^* A^{2t} V$ is positive selfadjoint operator for any $0 \leq t \leq 1/2$, therefore we can apply Prop.~\ref{prop:lop-prop} two times, obtaining
$$\nor{A^r V (B V^* A^s)} \leq \nor{(V^* A V)^{r} (B V^* A^s)} = \nor{((V^* A V)^{r} B) V^* A^s} \leq \nor{((V^* A V)^{r}B)(V^* A V)^{s}}.$$
\epr

\bp\label{prop:invAtoinvB}
Let $\hh$ be a separable Hilbert space, let $A, B$ two bounded self-adjoint positive linear operators and $\la > 0$. Then
$$ \nor{(A + \lambda I)^{-1/2}B^{1/2}} \leq \nor{(A + \lambda I)^{-1/2}(B + \la I)^{1/2}} \leq (1-\beta)^{-1/2}$$
when
$$ \beta = \la_{\max}\left[(B+\lambda I)^{-1/2}(B-A)(B+\lambda I)^{-1/2}\right] < 1.$$
\ep
\bpr Let $B_\la = B + \la I$. 
First of all we have,
\eqals{
\nor{(A + \lambda I)^{-1/2}B^{1/2}} &= \nor{(A + \lambda I)^{-1/2}B_\la^{1/2}B_\la^{-1/2}B^{1/2}} \\
&\leq \nor{(A + \lambda I)^{-1/2}B_\la^{1/2}}\nor{B_\la^{-1/2}B^{1/2}} \leq \nor{(A + \lambda I)^{-1/2}B_\la^{1/2}},
}
since $\nor{B_\la^{-1/2}B^{1/2}} = \sqrt{\frac{\nor{B}}{\nor{B} + \la}} \leq 1$.
Note that
\begin{align*}
(A+\lambda I)^{-1} & = \left[(B + \lambda I) - (B - A)\right]^{-1} \\
& = \left[B_\la^{1/2}\left(I - B_\la^{-1/2}(B - A)B_\la^{-1/2}\right)B_\la^{1/2}\right]^{-1}\\
& = B_\la^{-1/2}\left[I - B_\la^{-1/2}(B - A)B_\la^{-1/2}\right]^{-1}B_\la^{-1/2}.
\end{align*}
Now let $X = (I - B_\la^{-1/2}(B - A)B_\la^{-1/2})^{-1}$. We have that, 
\begin{align*}
\nor{(A + \lambda I)^{-1/2}B_\la^{1/2}}{} &= \nor{B^{1/2}(A + \lambda I)^{-1}B_\la^{1/2}}^{1/2}\\
&= \nor{B_\la^{1/2}B_\la^{-1/2}XB_\la^{-1/2}B_\la^{1/2}}^{1/2} = \nor{X}^{1/2},
\end{align*}
because $\nor{Z}{} = \nor{Z^*Z}{}^{1/2}$ for any bounded operator $Z$. 
Finally let $Y = B_\la^{-1/2}(B - A)B_\la^{-1/2}$ and assume that $\la_{\max}(Y) < 1$, then
$$\nor{X}{} = \nor{(I - Y)^{-1}}{} = (1 - \lambda_{\max}(Y))^{-1},$$
since $X = w(Y)$ with $w(\sigma) = (1-\sigma)^{-1}$ for $-\infty \leq \sigma < 1$, and $w$ is positive and monotonically increasing on the domain.  
\epr

\section{Tail bounds}
Let $\nor{\cdot}_{HS}$ denote the Hilbert-Schmidt norm.

\bp\label{prop:concentration-ClaC}
Let $v_1,\dots,v_n$ with $n \geq 1$, be independent and identically distributed random vectors on a separable Hilbert spaces $\hh$ such that $Q = \mathbb{E} \, v \otimes v$ exists, is trace class, and for any $\la > 0$ there exists a constant ${\cal N}_\infty(\la) < \infty$ such that $\scal{v}{(Q+\la I)^{-1}v} \leq {\cal N}_\infty(\la)$ almost everywhere. Let $Q_n = \frac{1}{n} \sum_{i=1}^{n} v_i \otimes v_i$ and take $0 < \lambda \leq \nor{Q}{}$.  Then for any $\delta \geq 0$, the following holds
$$
\nor{(Q+\lambda I)^{-1/2}(Q-Q_n)(Q+\lambda I)^{-1/2}}{} \leq \frac{2\beta(1 + {\cal N}_\infty(\la))}{3 n} + \sqrt{\frac{2\beta {\cal N}_\infty(\la)}{n}}$$
with probability $1-2\delta$. Here $\beta = \log \frac{4 \tr Q}{\lambda\delta}$. Moreover it holds that
$$
\la_{\max}\left((Q+\lambda I)^{-1/2}(Q-Q_n)(Q+\lambda I)^{-1/2}\right) \leq \frac{2\beta}{3 n} + \sqrt{\frac{2\beta {\cal N}_\infty(\la)}{n}}$$
with probability $1-\delta$. 
\ep
\bpr
Let $Q_\la = Q + \la I$.
Here we apply Prop.~\ref{prop:tail-operators} on the random variables $Z_i = M - Q_\la^{-1/2}v_i \otimes Q_\la^{-1/2}v_i$ with $M = Q_\la^{-1/2}QQ_\la^{-1/2}$ for $1 \leq i \leq n$. Note that the expectation of $Z_i$ is $0$.
The random vectors are bounded by
$$\nor{Q_\la^{-1/2}QQ_\la^{-1/2} - Q_\la^{-1/2}v_i \otimes Q_\la^{-1/2}v_i}{} \leq \scal{v}{Q_\la^{-1}v} + \nor{Q_\la^{-1/2}QQ_\la^{-1/2}}{} \leq {\cal N}_\infty(\la) + 1$$
and the second orded moment is
\begin{align*}
 \mathbb{E} (Z_1)^2 &= \mathbb{E} \;\;\scal{v_1}{Q_\la^{-1} v_1} \;Q_\la^{-1/2}v_1 \otimes Q_\la^{-1/2}v_1 \;\;\;-\;\;\; Q_\la^{-2}Q^2 \\
 & \leq {\cal N}_\infty(\la) \mathbb{E} Q_\la^{-1/2}v_1 \otimes Q_\la^{-1/2}v_1 = {\cal N}_\infty(\la) Q = S.
\end{align*}
Now we can apply Prop.~\ref{prop:tail-operators}. Now some considerations on $\beta$.
It is $\beta = \log \frac{4\tr S}{\nor{S}{}\delta} = \frac{4\tr Q_\la^{-1}Q}{\nor{Q_\la^{-1}Q}{}\delta}$, now $\tr Q_\la^{-1}Q \leq \frac{1}{\la} \tr Q$. We need a lowerbound for $\nor{Q_\la^{-1}Q}{} = \frac{\sigma_1}{\sigma_1 + \la}$ where $\sigma_1 = \nor{Q}$ is the biggest eigenvalue of $Q$, now $\la \leq \sigma_1$ thus $\frac{\ \tr Q}{\la\delta}$.

For the second bound of this proposition, the analysis remains the same except for $L$, indeed 
$$ \sup_{f \in \hh} \scal{f}{Z_1 f} = \sup_{f \in \hh} \scal{f}{Q_\la^{-1}Qf} - \scal{f}{Q_\la^{-1/2}v_i}^2 \leq \sup_{f \in \hh} \scal{f}{Q_\la^{-1}Qf} \leq 1.$$
\epr

\br\label{rem:ClC}
In Prop.~\ref{prop:concentration-ClaC}, let define $\kappa^2 = \inf_{\la>0} {\cal N}_\infty(\la)(\nor{Q} + \la)$. When $n \geq 405\kappa^2 \vee 67\kappa^2 \log \frac{\kappa^2}{2\delta}$ and $\frac{9\kappa^2}{n}\log\frac{n}{2\delta}\leq \la \leq \nor{Q}{}$ we have that 
$$\la_{\max}\left((Q+\lambda I)^{-1/2}(Q-Q_n)(Q+\lambda I)^{-1/2}\right) \leq \frac{1}{2},$$ with probability $1-\delta$, while it is less than $1/3$ with the same probability, if $\frac{19\kappa^2}{n}\log\frac{n}{4\delta}\leq \la \leq \nor{Q}{}$.
\er

\bp[Theorem 2 \cite{alaoui2014fast}. Approximation of matrix products.]\label{prop:conc-subsample}
Let $n, n$ be positive integers. Consider a matrix $A \in \R^{n \times n}$ and denote by $a_i$ the $i$-th column of $A$. Let $m \leq n$ and $I = \{i_1,\dots, i_m\}$ be a subset of $N = \{1,\dots,n\}$ formed by $m$ elements chosen randomly with replacement, according to a distribution that associates the probability $P(i)$ to the element $i \in N$.
Assume that there exists a $\beta \in (0,1]$ such that the probabilities $P(1),\dots,P(n)$ satisfy $ P(i) \geq \beta \frac{\nor{a_i}{}^2}{\tr AA^\top}$ for all $i \in N$.
For any $\delta > 0$ the following holds
$$\nor{AA^\top \;-\; \frac{1}{m}\sum_{i \in I} \frac{1}{P(i)}\, a_i a_i^\top}{} \leq \frac{2 L \log \frac{2n}{\delta}}{3m} + 
\sqrt{\frac{2L S\log\frac{2n}{\delta}}{m}}$$
with probability $1 - \delta$. Here $L = \nor{A}{}^2$ and $S = \frac{1}{\beta}\tr{AA^\top}$.
\ep

\bp[Bernstein's inequality for sum of random variables]\label{prop:tail-bern}
Let $x_1,\dots,x_n$ be a sequence of independent and identically distributed random variables on $\R$ with zero mean. If there exists an $L, S \in \R$ such that $x_1 \leq L$ almost everywhere and $\mathbb{E} x_1^2 \leq S$, then for any $\delta > 0$ the following holds with probability $1-\delta$:
$$\frac{1}{n} \sum_{i=1}^n x_i \leq \frac{2L\log\frac{1}{\delta}}{3n} + \sqrt{\frac{2S\log\frac{1}{\delta}}{n}}.$$
If there exists an $L' \geq |x_1|$ almost everywhere, then the same bound, computed with $L'$ instead of $L$, holds for the for the absolute value of the left hand side, with probability $1-2\delta$.
\ep
\bpr
It is a restatement of Theorem 3 of \cite{boucheron2004concentration}.
\epr

\bp[Bernstein's inequality for sum of random vectors]\label{prop:tail-vectors}
Let $z_1,\dots,z_n$ be a sequence of independent identically distributed random vectors on a separable Hilbert space $\hh$. Assume $\mu = \mathbb{E} z_1$ exists and let $\sigma, M \geq 0$ such that
$$ \mathbb{E} \nor{z_1 - \mu}_{\hh}^p \leq \frac{1}{2}p!\sigma^2 L^{p-2} $$
for all $p \geq 2$. Then for any $\tau \geq 0$:
$$\nor{\frac{1}{n}\sum_{i=1}^n z_i - \mu}_{\hh} \leq \frac{2 L \log\frac{2}{\delta}}{n} + \sqrt{\frac{2 \sigma^2\log\frac{2}{\delta}}{n}}$$
with probability greater or equal $1 - \delta$.
\ep
\bpr
restatement of Theorem 3.3.4 of \cite{yurinsky1995sums}.
\epr
\bp[Bernstein's inequality for sum of random operators]\label{prop:tail-operators}
Let $\hh$ be a separable Hilbert space and let $X_1,\dots,X_n$ be a sequence of independent and identically distributed self-adjoint positive random operators on $\hh$. Assume that there exists $\mathbb{E} X_1 = 0$ and $\la_{\max}(X_1) \leq L$ almost everywhere for some $L > 0$. Let $S$ be a positive operator such that $\mathbb{E} (X_1)^2 \leq S$. Then for any $\delta \geq 0$ the following holds
\begin{align*}
\la_{\max}\left(\frac{1}{n}\sum_{i=1}^n X_i\right) \leq \frac{2L\beta}{3n} + \sqrt{\frac{2\nor{S}{}\beta}{n}}
\end{align*}
with probability $1-\delta$. Here $\beta = \log\frac{2\tr S}{\nor{S}{}\delta}$. 

If there exists an $L'$ such that $L' \geq \nor{X_1}{}$ almost everywhere, then the same bound, computed with $L'$ instead of $L$, holds for the operatorial norm with probability $1-2\delta$.
\ep
\bpr
The theorem is a restatement of Theorem 7.3.1 of  \cite{tropp2012user} generalized to the separable Hilbert space case by means of the technique in Section 4 of \cite{minsker2011some}.
\epr
{\small
\putbib[nipsBib]
}
\end{bibunit}
\end{document}